\documentclass[final]{cvpr}

\usepackage{times}
\usepackage{epsfig}
\usepackage{graphicx}
\usepackage{amsmath}
\usepackage{amssymb}
\usepackage[dvipsnames]{xcolor}

\newcommand{\dsname}{TextSeg}
\newcommand{\modelfullname}{Text Refinement Network}
\newcommand{\modelname}{TexRNet}

\providecommand{\zz}[1]{}
\providecommand{\xx}[1]{}
\providecommand{\bp}[1]{}
\providecommand{\zw}[1]{}
\providecommand{\hs}[1]{}
\usepackage[normalem]{ulem}
\newcommand{\xxsub}[2]{#2}

\usepackage{booktabs}
\usepackage{array}
\newcolumntype{L}[1]{>{\raggedright\let\newline\\\arraybackslash\hspace{0pt}}m{#1}}
\newcolumntype{C}[1]{>{\centering\let\newline\\\arraybackslash\hspace{0pt}}m{#1}}
\newcolumntype{R}[1]{>{\raggedleft\let\newline\\\arraybackslash\hspace{0pt}}m{#1}}
\usepackage{tablefootnote}
\usepackage{caption} 
\usepackage{subcaption}
\usepackage{multirow}

\usepackage[pagebackref=true,breaklinks=true,colorlinks,bookmarks=false]{hyperref}

\def\cvprPaperID{5430} % *** Enter the CVPR Paper ID here
\def\confYear{CVPR 2021}

\begin{document}

%%%%%%%%% TITLE
\title{Rethinking Text Segmentation: \\A Novel Dataset and A Text-Specific Refinement Approach}

\author{Xingqian Xu$^1$, Zhifei Zhang$^2$, Zhaowen Wang$^2$, Brian Price$^2$, Zhonghao Wang$^1$, Humphrey Shi$^{1,3}$\\
\\
$^1$UIUC, $^2$Adobe, $^3$University of Oregon}

\maketitle

%%%%%%%%% ABSTRACT
\begin{abstract}

Text segmentation is a prerequisite in many real-world text-related tasks, \eg, text style transfer, and scene text removal. However, facing the lack of high-quality datasets and dedicated investigations, this critical prerequisite has been left as an assumption in many works, and has been largely overlooked by current research. To bridge this gap, we proposed \textbf{\dsname{}}, a large-scale fine-annotated text dataset with six types of annotations: word- and character-wise bounding polygons, masks and transcriptions. We also introduce \modelfullname{} (\textbf{\modelname{}}), a novel text segmentation approach that adapts to the unique properties of text, \eg non-convex boundary, diverse texture, etc., which often impose burdens on traditional segmentation models. In our TexRNet, we propose text specific network designs to address such challenges, including key features pooling and attention-based similarity checking. We also introduce trimap and discriminator losses that show significant improvement on text segmentation. Extensive experiments are carried out on both our \dsname{} dataset and other existing datasets. We demonstrate that \modelname{} consistently improves text segmentation performance by nearly 2\% compared to other state-of-the-art segmentation methods. Our dataset and code will be made available at \href{https://github.com/SHI-Labs/Rethinking-Text-Segmentation}{https://github.com/SHI-Labs/Rethinking-Text-Segmentation}.

\end{abstract}

\vspace{-0.2cm}
\section{Introduction}
Text segmentation is the foundation of many text-related computer vision tasks. It has been studied for decades as one of the major research directions in computer vision, and it continuously plays an important role in many  applications~\cite{tstyle_mcgn, tstyle_yang, shapemgan, deep_image_prior, im_edit}. Meanwhile, the rapid advances of deep neural nets in recent years promoted all sorts of new text related research topics, as well as new vision challenges on text. Smart applications, such as font style transfer, scene text removal, and interactive text image editing, require effective text segmentation approaches ahead to accurately parse text from complex scenes. Without any doubt, text segmentation is critical for industrial usages because it could upgrade the traditional text processing tools to be more intelligent and automatic, relaxing tedious efforts on manually specifying text regions.

\begin{figure}[t!]
    \centering
    \includegraphics[width=0.48\textwidth]{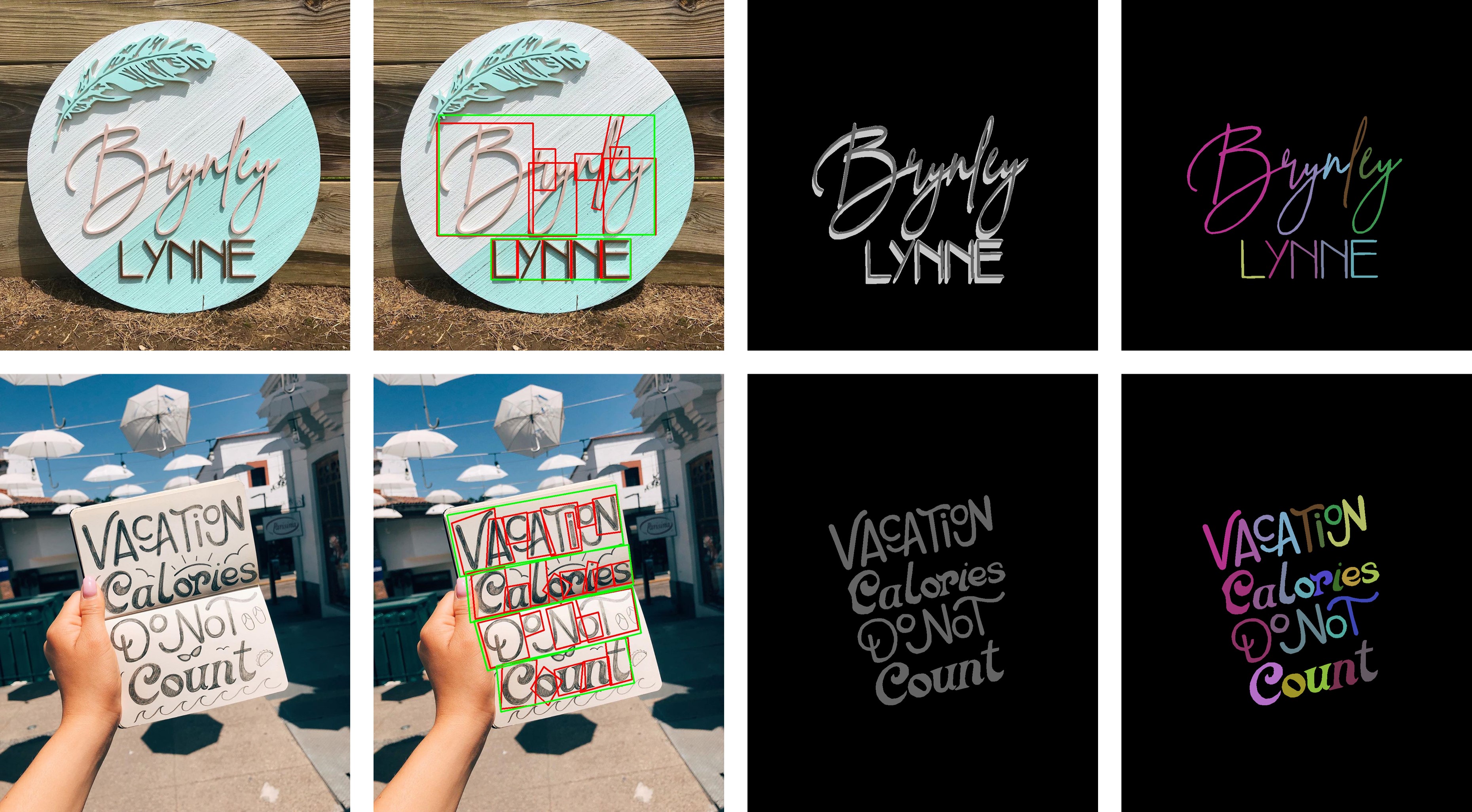}
    \caption{Example images and annotations from the proposed \dsname{} dataset. From left to right are images, word and character bounding polygons, pixel-level word (dark gray) and word-effect (light gray) masks, and pixel-level character masks.}
\label{fig:image_anno}
\vspace{-0.55cm}
\end{figure}

\begin{figure*}[t!]
    \centering
    \includegraphics[width=0.99\textwidth]{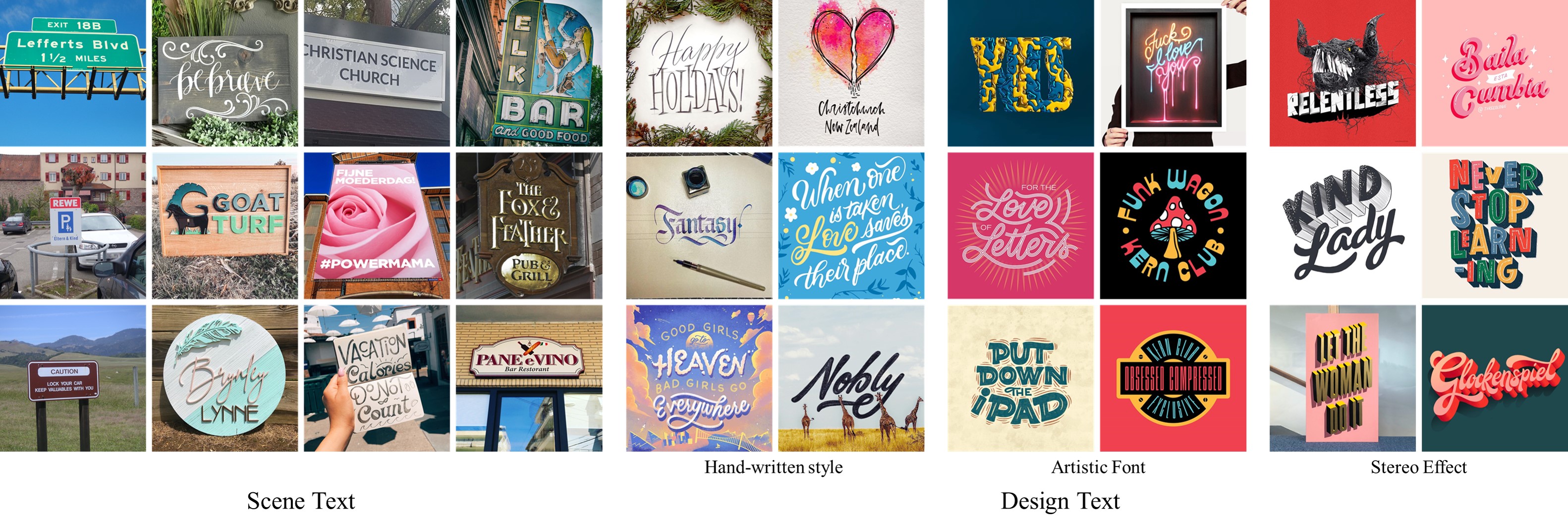}
    \vspace{-0.3cm}
    \caption{Images examples from the proposed \dsname{} dataset. The left four columns show scene text that dominantly presents in existing text segmentation datasets, and the rest columns are design text w/ or w/o text effects, which distinguishes \dsname{} from all the other related datasets.}
\label{fig:image_only}
\vspace{-0.3cm}
\end{figure*}

However, modern text segmentation has been left behind in both datasets and methods. The latest public text segmentation challenge was in 2013-2015 hosted by ICDAR~\cite{icdar13}. Since then, three datasets: Total-Text~\cite{totaltext}, COCO\_TS~\cite{cocots} and MLT\_S~\cite{mlts} were introduced. However, Total-Text is limited in scale, and the labeling quality in COCO\_TS and MLT\_S needs further improvement (Figure~\ref{fig:anno_compare}). Moreover, all the three datasets contain only common scene text, discouraging text in other visual conditions, \eg, artistic design and text effects. As a result, these datasets do not meet the standards of modern research such as large-scale and fine-annotated. Thus, we purpose a new text segmentation dataset: \textbf{\dsname{}} that collects images from a wider range of sources, including both scene and design text, and with a richer set of accurate annotations. This dataset would lead to further advancements in text segmentation research.

Additionally, text segmentation algorithms and methods in recent years fall behind from other research topics, partially due to the lack of a proper dataset. Unlike the rapid advances in other segmentation research, only a few studies~\cite{cnn_texseg_ccnn, mlts, cnn_texseg_nstd} have brought new ideas to text segmentation. Meanwhile, these studies did not provide an intuitive comparison with modern SOTA segmentation approaches, and thus were unable to demonstrate their advantages over other techniques. As aforementioned, effective text segmentation models are valuable in applications. With our strong motivation in bridging this gap, we propose \modelfullname{} (\textbf{\modelname{}}), and we thoroughly exam its performance on five text segmentation datasets including the proposed \dsname{} dataset. The details of our design principles and our network structure are given in Session~\ref{sec:approach}, and experiments and ablations studies are shown in Session~\ref{sec:experiment}.

In summary, the main contributions of this paper are in three-folds: 
\vspace{-0.2cm}
\begin{itemize}
    \item[$\bullet$] We introduce a new large-scale fine-annotated text segmentation dataset, \dsname{}, consisting of 4,024 text images including scene text and design text with various artistic effects. For each image, \dsname{} provides six types of annotations, \ie, word- and character-wise quadrilateral bounding polygons, pixel-level masks, and transcriptions. Our dataset surpasses previous works on these aspects: 1) more diverse text fonts/styles from diverse sources/collections, 2) more comprehensive annotations, and 3) more accurate segmentation masks. 

    \vspace{-0.1cm}
    \item[$\bullet$] We provide a new text segmentation approach, \modelfullname{} (\modelname{}), aiming to solve the unique challenges from text segmentation. We design effective network modules (\ie, key features pooling and attention-based similarity checking) and losses (\ie, trimap loss and glyph discriminator) to tackle those challenges, \eg, diverse texture and arbitrary scales/shapes. 
    
    \vspace{-0.1cm}
    \item[$\bullet$] Exhaustive experiments are conducted to demonstrate the effectiveness of the proposed \modelname{}, which outperforms SOTA not only on our \dsname{}, but also on another four representative datasets. In addition, we give prospects for downstream applications that could significantly benefit from the text segmentation.  
\end{itemize}

\vspace{-0.2cm}
\section{Related Work}
\label{sec:2}

% This section gives an overview on datasets and methods that closely related to ours. The first part is an introduction on methods and models created for semantic and instance segmentation. Next, we summarize past text segmentation works and highlight . In the end, we go review 

\subsection{Segmentation in Modern Research}

Semantic and instance segmentation are popular tasks for modern research. In semantic segmentation, pixels are categorized into a fixed set of labels. Datasets such as PASCAL VOC~\cite{pascal_voc}, Cityscapes~\cite{cityscapes}, COCO~\cite{coco}, and ADE20K~\cite{ade20k} are frequently used in this task. Traditional graph models,\eg, MRF~\cite{mrf} and CRF~\cite{crf}, predict segments by exploring inter-pixel relationship. After CNNs became popular~\cite{cnn}, numerous deep models were proposed using dilated convolutions~\cite{pspnet, deeplabv3, deeplabv3+, hrnet}, encoder-decoder structures~\cite{unet, pspnet, deeplabv3+, fpn}, and attention modules~\cite{nonlocal, attn, danet}. Instance segmentation methods predict distinct pixel labels for each object instance. These methods can be roughly categorized into top-down approaches~\cite{sds, fcis, maskrcnn, panet, upsnet, pfpn} and bottom-up approaches~\cite{dwt, ssap, gmis, daffnet, josecb}. Top-down approaches are two-stage methods that first locate object bounding boxes and then segment object masks within those boxes. Bottom-up approaches locate keypoints~\cite{deeperlab, josecb} and find edges and affinities~\cite{ssap, gmis, daffnet, dwt} to assist the segmentation process. 

\subsection{Text Segmentation}

Early methods frequently used thresholding~\cite{tra_texseg_thr1, tra_texseg_thr2} for segmentation particularly on document text images. Yet such methods cannot produce satisfactory results on scene text images with complex colors and textures. Other approaches used low-level features~\cite{tra_texseg_edge, tra_texseg_colr, tra_texseg_seed} and Markov Random Field (MRF)~\cite{mrf_texseg} to bipartite scene text images. In~\cite{tra_texseg_edge}, text features created from edge density/orientation were fed into an multiscale edge-based extraction algorithm for segmentation. In~\cite{tra_texseg_colr}, a two-stage method was introduced in which foreground color distribution from stage one was used to refine the result for stage two. In~\cite{tra_texseg_seed}, seed points of both text and background were extracted from low-level features and were later used in segmentation. Inspired by MRF, \cite{mrf_texseg} formulated pixels as random variables in a graph model, and then graph-cut this model with two pre-selected seeds. In recent years, several deep learning methods~\cite{cnn_texseg_ccnn, cnn_texseg_nstd, mlts} were proposed for text segmentation. The method proposed by~\cite{cnn_texseg_ccnn} is a three-stage CNN-based model, in which candidate text regions were detected, refined, and filtered in those stages correspondingly. Another method SMANet was jointly proposed with the dataset MLT\_S in~\cite{mlts}. They adopted the encoder-decoder structure from PSPNet~\cite{pspnet}, and created a new multiscale attention module for accurate text segmentation. 

\zz{Any drawbacks from previous deep learning based works?}\xx{will do it later}

\subsection{Text Dataset}

Spotlight datasets motivate researchers to invent effective methods to tackle computer vision problems. For example, the MNIST dataset of handwritten digits~\cite{mnist} illustrated the effectiveness of a set of classical algorithms, \eg, KNN~\cite{knn}, PCA~\cite{pca}, SVM~\cite{svm}, etc. In recent years, the huge success of deep learning inspires researchers to create more challenging datasets to push forward the vision research front. Many text datasets are created for OCR purpose like CUTE80~\cite{cute80}, MSRA-TD500~\cite{msratd500}, ICDARs~\cite{icdar15, icdar13, icdar17mlt}, COCO-Text~\cite{cocotext}, and Total-Text~\cite{totaltext}, which are scene text datasets with word-level bounding boxes. Other datasets such as Synth90K~\cite{syn90k} and SynthText~\cite{syntex} are synthetic text dataset for recognition and detection. Among these dataset, ICDAR13~\cite{icdar13} and Total-Text~\cite{totaltext} provide pixel-level labels for text segmentation. Recently, Bonechi \etal introduced segmentation labels to COCO-Text and ICDAR17 MLT, forming up two new text segmentation datasets COCO\_TS~\cite{cocots} and MLT\_S~\cite{mlts}. 
In general, ICDAR13 and Total-Text are relatively smaller sets, and COCO\_TS and MLT\_S are of larger scale but their labeling quality is not precise.

\section{\modelfullname{}}
\label{sec:approach}

\begin{figure*}[t!]
    \centering
    \includegraphics[width=0.95\textwidth]{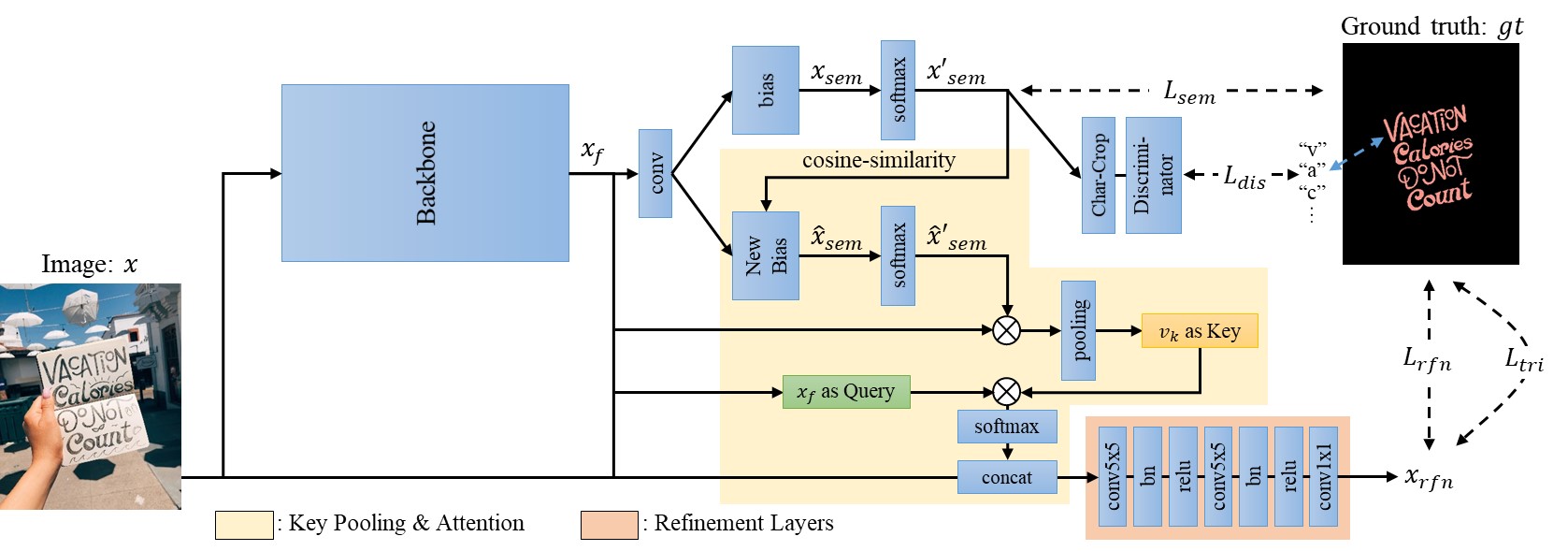}
    \caption{Overview of the proposed \modelname{}. Most parts of the network are well-explained in Session~\ref{subsec:texrnet}. Besides, "Char-Crop" is the modules that help cropping characters for classifier. It requires ground truth character bounding boxes as input, and therefore will only be available if those boxes are provided. During inference time, neither "Char-Crop" nor "Classifier" need to be loaded, and $x_{rfn}$ will be the model's final output.}
\label{fig:network}
\vspace{-0.2cm}
\end{figure*}

We propose a \xxsub{novel deep module}{new approach}, namely \modelfullname{} (\modelname{}), which specifically targets text segmentation. Since text segmentation is intrinsically similar to modern semantic segmentation, these related state-of-the-art methods can be leveraged to provide the base for our proposed \modelname{}. Figure~\ref{fig:network} overviews the pipeline of \modelname{}, which consists of two components: 1) a backbone, \eg, DeeplabV3+~\cite{deeplabv3+} or HRNet~\cite{hrnet}, and 2) the key features pooling and attention module that refines the backbone for the text domain. The design of the latter module is inspired by the uniqueness of text segmentation, and the principles will be discussed in Section~\ref{subsec:design_principle}. The network structure and corresponding loss functions will be detailed in Sections~\ref{subsec:texrnet} and \ref{subsec:losses}, respectively.  

\subsection{Design Principle}
\label{subsec:design_principle}

Multiple unique challenges distinguish text segmentation from \xxsub{traditional}{modern} semantic segmentation, thus motivating specific designs for text segmentation. 
In semantic segmentation, common objects, \eg, trees, sky, cars, etc., tend to share texture across different scenes. In text segmentation, however, the text texture may be extremely diverse across different words although it could be homogeneous inside each word. 
To accommodate larger texture diversity, \modelname{} dynamically activates low-confidence areas according to their global similarity to high-confidence regions, \ie, the yellow block in Figure~\ref{fig:network}, which aims to adaptively find similar textures in the same scene while relaxing the model from ``remembering'' those diverse textures. 

Another challenge of text segmentation is the arbitrarily scaled text. The commonly adopted convolutional layers in semantic segmentation would limit the receptive field, reducing adaptiveness to diverse scale and aspect ratio. 
To achieve higher adaptiveness to scale, 
%\xxsub{we design a query-key attention mechanism~\cite{attention}, \ie, the green and yellow blocks in Figure~\ref{fig:network}, which enforces global attention and interchange across the entire image. The experimental evaluation (Sec.~\ref{sec:experiment}) has demonstrated the effectiveness of both dynamically query-key attention module.}{}
we adopt the popular non-local concept~\cite{nonlocal, attn}, from which we use dot product and softmax to enforce attention on similar texture across the entire image.  

\subsection{Network Structure}
\label{subsec:texrnet}

As aforementioned, the backbone can employ an arbitrary semantic segmentation network. Here, we choose two representative works, \ie, ResNet101-DeeplabV3+~\cite{deeplabv3+} and HRNetV2-W48~\cite{hrnet}, because they are the milestone and state-of-the-art in semantic segmentation, respectively. The rest of this section will focus on the new designs of \modelname{}, \ie, the yellow block in Figure~\ref{fig:network}, which is the key to boosting the performance of text segmentation.

Assume an input image $x\in \mathbb{R}^{H\times W \times 3}$, where $H$ and $W$ denote image height and width, respectively. The feature maps extracted from the backbone is $x_f$. The remainder of the proposed \modelname{} could be described in the following three sequential components.

% \begin{enumerate}
% 	\item 
    \vspace{0.2cm}
	\textbf{Initial Prediction:}
	Similar to most traditional segmentation models, the feature map $x_f$ is mapped to the semantic map $x_{sem}$ through a convolutional layer (the kernel size is $1\times 1$) with bias. After the softmax layer, $x_{sem}$ becomes the initial segmentation prediction $x'_{sem}$, which can be supervised by ground truth labels as the following.
    \begin{align}
        \mathcal{L}_{sem} = \text{CrossEntropy}(x'_{sem}, x_{gt}), 
        \label{eq:sem}
    \end{align}
	where $x'_{sem} = \text{Softmax}(x_{sem})$, and $x_{gt}$ indicates the ground truth label.

    % \bp{I think that at the beginning of this next section we should remind the reader why we are doing this.  We can say something like "Because text does not have a standard texture that can be learned during training, the network must determine that text texture during inference by identifying the shapes that correspond to glyphs.  However, this often results in the edges or tips of text glyphs having low confidence.  To address this, we identify the features that correspond to the text in high confidence regions and use that to better segment the low confidence regions."  Something like this to make sure the reader follows why this next part is useful.}
    % \xx{Added few new sentenses on that.}
	
% 	\item 
    \vspace{0.2cm}
	\textbf{Key Features Pooling:}
	Because text does not have a standard texture that can be learned during training, the network must determine that text texture during inference. Specifically, the network should revise low-confidence regions if they share similar texture with high-confidence regions of the same class. 
	To achieve this goal, we need to pool the key feature vector from high-confidence regions for each class $i\in C$ to summarize the global visual property of that class. In our case, $|C|=2$, corresponding to text and background. More specifically, we conduct a modified cosine-similarity on the initial prediction $x'_{sem}$ and use its output as new biases to transform $x'_{sem}$ into $\hat{x}'_{sem}$ which is the weight map for key pooling. The cosine-similarity is written in Eq.~\ref{eq:cossim}, assuming $x'_{sem} \in \mathbb{R}^{c\times n}$, where $c=|C|$ denotes the number of classes, and $n$ is the number of pixels in a single channel. 
	\begin{equation}
	\begin{aligned}
    \text{CosSim}(x'_{sem}) &= X \in \mathbb{R}^{c\times c}, \\
    	    X_{ij} &= \begin{cases}
    	        \displaystyle\frac{x_i x_j^T}{||x_i||\cdot||x_j||}, &i\ne j\\
    	        0, &i=j
    	    \end{cases}, \\
    	    x_i &= {x'}_{sem}^{(i,)} \in \mathbb{R}^{1\times n}, \; i=1, \cdots, c,
	\end{aligned}
	\label{eq:cossim}
	\end{equation}
	where $\text{CosSim}(\cdot)$ denotes the modified cosine-similarity function, and ${x'}_{sem}^{(i,)}$ denotes the $i$th channel of $x'_{sem}$, \ie, the predicted score map on class $i$. From our empirical study, the cosine-similarity value $X_{ij}$ indicates the ambiguity between prediction on classes $i$ and $j$. For example, when $X_{ij}$ is close to 1, pixels are activated similarly in both ${x'}_{sem}^{(i,)}$ and ${x'}_{sem}^{(j,)}$, and thus cannot be trusted. Therefore, we use zero bias on class $i$ and use biases in proportional to $X_{ij}$ on class $j\ne i$ equivalent to decrease the confidence scores on class $i$. Those regions remains high-activated in class $i$ are then confidence enough for the key pooling. The final key pooling is a normalized weighted sum between the weight map $\hat{x}'_{sem}$ and feature map $x_f$: 
    \begin{align}
		v_i=\frac{
		    x_f \cdot 
		    \left( \hat{x}'^{(i,)}_{sem} \right)^T
		} 
		{\big\|\hat{x}'^{(i,)}_{sem}\big\|}, \; 
		v=\begin{bmatrix}v_i, \cdots \end{bmatrix}, \;
		i=1, \cdots, c,
		\label{eq:v}
	\end{align} 
	where $x_f\in\mathbb{R}^{m\times n}$ denotes the feature map with $m$ channels and $n$ pixels in each channel, $v_i\in\mathbb{R}^{m\times 1}$ denotes the pooled vector for class $i$, and $v\in\mathbb{R}^{m\times c}$ is the concatenated matrix from $v_i$.

% 	\item 
    \vspace{0.2cm}
	\textbf{Attention-based Similarity Checking:}
	We then adopt an attention layer, which uses $v$ as key and $x_f$ as query, and computes the query-key similarity $x_{att}$ through dot-product followed by softmax:
	\begin{align}
		x_{att} = \text{Softmax}(v^T \cdot x_f), \; x_{att}\in\mathbb{R}^{c\times n}.
		\label{eq:xatt}
	\end{align}
    The $x_{att}$ will activate those text regions that may be ignored due to low-confidence in the initial prediction $x'_{sem}$. Then, we fuse $x_{att}$ with the input image $x$ and backbone feature $x_f$ into our refined result $x_{rfn}$ through several extra convolutional layers (orange block in Figure~\ref{fig:network}). Note that our attention layer differs from the traditional query-key-value attention~\cite{attn} in several ways. Traditional attention requires identical matrix dimensions on query and key, while our approach uses a key $v$ that is significant smaller than the query $x_f$. Also, traditional attention fuses value and attention through dot product, while ours fuses $x_{att}$ with other features through a deep model. 
    The final output $x_{rfn}$ is supervised by the ground truth as shown in the following.
    \begin{align}
    	\mathcal{L}_{rfn} = \text{CrossEntropy}(x_{rfn}, x_{gt}). 
    	\label{eq:rfn}
    \end{align} 

% \end{enumerate}

\subsection{Trimap Loss and Glyph Discriminator}
\label{subsec:losses}

% Besides the network structure, two additional losses are introduced in our training for better segmentation results. One is the trimap cross-entropy loss $L_{tri}$. Inspired by~\cite{trimapiou}, we locate all text boundary pixels and find another cross-entropy loss only on these pixels. One can also think of it as a weight cross-entropy loss where boundary pixels are weighted as 1 and all other pixels as zero. $L_{tri}$ is calculated using the refined result $x_{rfn}$ (attached to the refinement branch). The other loss is the classifier loss $L_{cls}$. Since ground truth character bounding boxes are given in \dsname{}, we can pre-train a classifier for character recognition. When we train \modelname{}, we crop the predicted maps using character bounding boxes and pass them into the weight-freeze classifier. The classifier will then gives a classification loss showing how recognizable these cropped characters predicted maps are. Unlike $L_{tri}$, $L_{cls}$ is calculated on $x_{sem}$. In our experiment, we notice that the improvement by using $L_{tri}$ and $L_{cls}$ may be canceled out if they work on the same predicted map. This is because $L_{rfn}$ focuses on boundary accuracy while $L_{cls}$ focuses on the body structure of the text. Also, $L_{rfn}$ has an immediate effect over model results while $L_{cls}$ should help to shape the text masks on non-immediate results. 

% The final loss of \modelname{} is the weighted sum of all losses. Unless other specifications, we use weight 1, 0.5, 0.5 and 0.1 on $L_{sem}$, $L_{rfn}$, $L_{tri}$ and $L_{cls}$. 

Since human vision is sensitive to text boundaries, segmentation accuracy along the text boundary is of central importance. In addition, text typically has relatively high contrast between the foreground and background to make it more readable. Therefore, a loss function that focuses on the boundary would further improve the precision of text segmentation. Inspired by~\cite{trimapiou}, we proposed the trimap loss as expressed as follows,
% \begin{align}
% 	\mathcal{L}_{tri} = \text{CrossEntropy}\left( x_{rfn}\cdot\text{Lap}_\sigma (x_{gt}), \text{Lap}_\sigma (x_{gt})  \right),
% \end{align}
\begin{equation}
    \begin{gathered}
    	\mathcal{L}_{tri} = \text{WCE}\left(
    	    x_{rfn}, x_{gt}, w_{tri} \right),\\
    	\text{WCE}(x,y,w) = -\frac{\sum_{j=1}^{n} w_j \sum_{i=1}^{c} x_{i,j} \log(y_{i,j})}{\sum_{j=1}^{n}w_j} 
    \end{gathered}
\end{equation}
% \xxsub{where $\text{Lap}_\sigma(\cdot)$ denotes the Laplacian filter with boundary relax paramterized by $\sigma$. Intuitively, $\sigma$ controls the boundary width, \ie, pixels with the distance to the boundary smaller than $\sigma$ will be involved in $\mathcal{L}_{tri}$. For instance,  if $\sigma=0$, only boundary is considered; if $\sigma=5$, areas within 5 pixels along the boundary are counted. Typically, $\text{Lap}_\sigma(\cdot)$ provides a binary mask, where 1 indicates the boundary area and 0 otherwise.}{}
where $w_{tri}$ is the binary map with value 1 on text boundaries and 0 elsewhere, and WCE$(x,y,w)$ is cross-entropy between $x$ and $y$ weighted by the spatial map $w$. 

% \zz{no clear on ``normalize-weighted CrossEntropy''}
% \xx{New formula, see whether it is clear this time.}

Another unique attribute of text is its readable nature, \ie, the segments of glyphs should be perceptually recognizable. Given that the partial segmentation of a glyph diminishes its readability, we train a glyph discriminator to improve the readability of text segments. It is worth noting that the glyph discriminator also improves the evaluation score \xxsub{under traditional segmentation metrics} as shown in the experimental evaluation. More specifically, we pre-train a classifier for character recognition given the ground-truth character bounding boxes in the training set (the proposed dataset \dsname{} provides these annotations). In our case, there are 37 classes, \ie, 26 letters, 10 digits, and misc. During the training of \modelname{}, the pre-trained classifier is frozen and applied to the initial prediction $x'_{sem}$, serving as the glyph discriminator. As illustrated in Figure~\ref{fig:network}, $x'_{sem}$ is cropped into patches according to the character locations, and then fed into the discriminator to obtain the discriminator loss $\mathcal{L}_{dis}$, which indicates whether and how these patches are recognizable. 

Unlike $\mathcal{L}_{tri}$ that operates on $x_{rfn}$, the glyph discriminator is applied on the initial prediction $x'_{sem}$ for mainly two reasons: 1) $\mathcal{L}_{tri}$ focuses on boundary accuracy while $\mathcal{L}_{dis}$ focuses on the body structure of the text, which ``distracts'' each other if they are applied on the same prediction map. Our empirical studies also show that the improvements from  $\mathcal{L}_{tri}$ and $\mathcal{L}_{dis}$ would be diminished if they work together on the same output, which aligns with our analysis. 2) $\mathcal{L}_{tri}$ can directly impact the performance so it oversees the model's final output $x_{rfn}$, while $\mathcal{L}_{dis}$ reinforces the deep perception on text thus it can be placed on earlier layers.
Above all, the final loss of \modelname{} will be 
\begin{align}
	\mathcal{L} =  \mathcal{L}_{sem} + \alpha \mathcal{L}_{rfn} + \beta \mathcal{L}_{tri} + \gamma \mathcal{L}_{dis},
\end{align}
where $\alpha$, $\beta$, and $\gamma$ are weights from 0 to 1. In the following experiments,  $\alpha=0.5$, $\beta=0.5$, and $\gamma=0.1$.

\section{The New Dataset \dsname{}}
\begin{table*}
\centering
\resizebox{0.99\textwidth}{!}{
    \begin{tabular}{
            R{3.5cm}|C{2cm}|C{2cm}|C{2.8cm}|C{1.3cm}|C{1.3cm}|C{2.8cm}|C{2cm}}
        \toprule
            \multicolumn{1}{c|}{\multirow{2}{*}{Dataset}} &  
                \multirow{2}{*}{\# Images} & 
                Approx. &
                \multirow{2}{*}{Text Type} &
                \multicolumn{2}{c|}{\# Bounding} &
                Word-level &
                \# Character\\
            \multicolumn{1}{c|}{} &  
                \multirow{2}{*}{} & 
                Image Size &
                \multirow{2}{*}{} & 
                \multicolumn{2}{c|}{Polygons} & 
                Masks & 
                Masks \\

        \midrule
            \multicolumn{1}{l|}{\textit{Scene Text Segmentation}} 
                &
                &
                &
                & \# Word
                & \# Char
                &
                & \\
                  ICDAR13 FST \cite{icdar13} 
                & 462
                & $1000 \times 700$
                & Scene 
                & 1,944
                & 6,620 
                & Word
                & 4,786 \\
                  COCO\_TS$^{\dagger}$ \cite{cocots} 
                & 14,690
                & $600 \times 480$
                & Scene 
                & 139,034
                & -- 
                & Word 
                & -- \\
                  MLT\_S$^{\dagger}$ \cite{mlts} 
                & 6,896
                & $1800 \times 1400$
                & Scene 
                & 30,691
                & -- 
                & Word 
                & -- \\
                  Total-Text \cite{totaltext} 
                & 1,555
                & $800 \times 700$
                & Scene 
                & 9,330
                & -- 
                & Word 
                & -- \\
                  \textbf{\dsname{} (Ours)} 
                & 4,024
                & $1000 \times 800$
                & Scene + Design
                & 15,691
                & 73,790
                & Word, Word-Effect
                & 72,254 \\
        \bottomrule
    \end{tabular}
}
\resizebox{0.99\textwidth}{!}{
    \begin{tabular}{L{0.2cm}L{27cm}}
        $^{\dagger}$ 
        & The 14,690 images in COCO\_TS is a subset of the totally 53,686 images in COCO-Text~\cite{cocotext}. Similarly, the 6,898 images in MLT\_S is a subset of the 10,000 images in ICDAR17 MLT~\cite{icdar17mlt}. Thus, their word bounding polygons can be directly extracted from their parent datasets.
        % \\
        % $^{\ddagger}$ 
        % & Pixel-level semantic labels are not given in ICDAR13 FST but can be inferred from their instance labels. 
    \end{tabular}
}
\vspace{0.2cm}
\caption{
    % This table shows the statistics comparison between numerous text segmentation datasets. If a dataset doesn't provide a certain type of annotation, we will place a "-" on that entry. 
    Statistical comparison between \dsname{} and other datasets for text segmentation. The ``--'' marker indicates absence of the corresponding annotation in a dataset.
}
\label{table:ds_compare}
\vspace{-0.1cm}
\end{table*}

\label{sec:dataset}

% To accommodate the rapid advances of the text vision research, we introduce \dsname{} which is a multi-purpose text dataset focused on segmentation. It contains 4024 high-resolution text images collected from a wide range of vision conditions. Meanwhile, \dsname{} is equipped with 4 types of annotations that help promoting research not only in segmentation but in OCR, style transfer, inpainting, etc. In this session, we will describe \dsname{}'s images, annotations and statistic with details. We will also promote \dsname{}'s high-quality by comparing its annotations with other text datasets. 

%\xxsub{As \xxsub{scene}{} text in the real world is extremely diverse in terms of vision condition (\eg, shadow, occlusion, and transparency), font/style (\eg, special design and 3D effect), texture/boundary, etc., it still lacks such a diverse and large-scale \xxsub{scene-text}{} dataset in the academic community. }
As text in the real world is extremely diverse,
to bridge text segmentation to the real world and accommodate the rapid advances of the text vision research, we propose a new dataset \dsname{}, which is a multi-purpose text dataset focused on but not limited to segmentation. 
%\xxsub{It consists of 4,024 high-resolution scene \xxsub{}{and design} text images collected from a wide range of vision conditions. Meanwhile, \dsname{} is equipped with six types of annotations, \ie, word- and character-wise bounding boxes, masks, and \xxsub{translations}{transcriptions} \bp{we should probably say ``transcriptions''}\xx{all changed}, which would promote research not only in text segmentation but also in text detection and recognition, text style/font transfer, text inpainting, etc. This section will first detail the proposed \dsname{} in terms of image collection, annotations, and statistical analysis, then the advantages of \dsname{} over existing datasets will be demonstrated at last. Four representative datasets for text segmentation are adopted in the comparison, \ie, ICDAR13 Focused Scene Text (FST)~\cite{icdar13}, \xxsub{ICDAR17 Multi-Lingual Scene Text (MLT\_S)~\cite{icdar17mlt}}{MLT\_S~\cite{mlts}}, COCO\_TS~\cite{cocots}, and Total-Text~\cite{totaltext}. }{}
% \xx{ICDAR17 MLT and MLT\_S are different.}

\subsection{Image Collection}

% The source of the images in \dsname{} can be roughly categorized into two types. One is the scene text in which the context in images follow physical and stereo rules. The other is the design text where physical and stereo rules are subject to artistic effect. Examples of scene and design text are shown in Figure~\ref{fig:image_only}.

% Unlike previous text datasets, text styles in \dsname{} are unrestricted. Text captured with digital, handwritten and artistic styles are all included. Meanwhile, the orderings of text can be horizontal, vertical, curved, discrete, etc. We believe that these unconstrained text images can help creating methods that are robust towards all kinds of real-world scenarios. The arbitrary styles and orderings of \dsname{} also introduce new challenges for previous text models. 

% Unlike the unrestricted styles and orderings of text, languages in \dsname{} are European languages and are mostly English. Therefore, almost all characters in \dsname{} are 26 letters, 10 digits and punctuation. Similar setting can also be found in Total-Text, COCO-Text, ICDAR13, etc. 

The 4,024 images in \dsname{} are collected from posters, greeting cards, covers, logos, road signs, billboards, digital designs, handwriting, etc. The diverse image sources could be roughly divided into two text types: 1) scene text, \eg, road signs and billboards, and 2) design text, \eg, artistic text on poster designs. Figure~\ref{fig:image_only} shows examples of the two types. 
Existing text related datasets tend to focus on scene text, while \dsname{} balances the two text types to achieve a more real-world and diverse dataset. In addition, rather than focusing on text lines, the proposed \dsname{} includes a large amount of stylish text. Sharing the language setting from those representative text segmentation datasets, the proposed \dsname{} mainly focuses on English (\ie, case-sensitive alphabet, numbers, and punctuation).% but contains a few non-English Latin characters.

\subsection{Annotations}

\dsname{} provides more comprehensive annotations as compared to existing datasets. More specifically, \dsname{} has annotated the smallest quadrilateral, pixel-level mask, and transcription for every single word and character. Besides, text effects, \eg, shadow, 3D, halo, etc., are annotated in \dsname{}, which distinguishes text from traditional objects and significantly affects text segmentation. To the best of our knowledge, the proposed \dsname{} is the only dataset with such comprehensive annotation for text segmentation. 

\textbf{Smallest Quadrilaterals} are annotated to tightly bound words, characters, and punctuation. These quadrilaterals are recorded in the image coordinate (\ie, top-left origin, $x$ axis is horizontally right, and $y$ axis is vertically down), and the four vertices are ordered clockwise starting from the top-left corner in the natural reading direction. A smallest quadrilateral tightly bounds a word or character as shown in Figure~\ref{fig:image_anno}. In certain cases like blurry text or long strokes, the quadrilaterals would cover the core area of the text by ignoring the ambiguous boundary or decorative strokes.

\textbf{Pixel-level Masks} consists of word masks, character masks, and word-effect masks. The word mask could be considered as a subset of word-effect mask since a word mask only labels the word surface without the effects like shadow and decoration while the effect mask will cover the word and all effects from the word. Similar to word masks, the character masks label character surfaces without those effects. Borrowing the concept from \xxsub{traditional}{modern} segmentation, word masks enable semantic segmentation, and character masks allow instance segmentation. For character masks, the most challenging cases are handwriting and artistic styles, where there are no clear boundaries between characters, thus the criterion is to keep all masks perceptually recognizable.

\begin{figure}[t!]
\centering
    \begin{subfigure}{.23\textwidth}
        \centering
        \includegraphics[width=.99\linewidth]{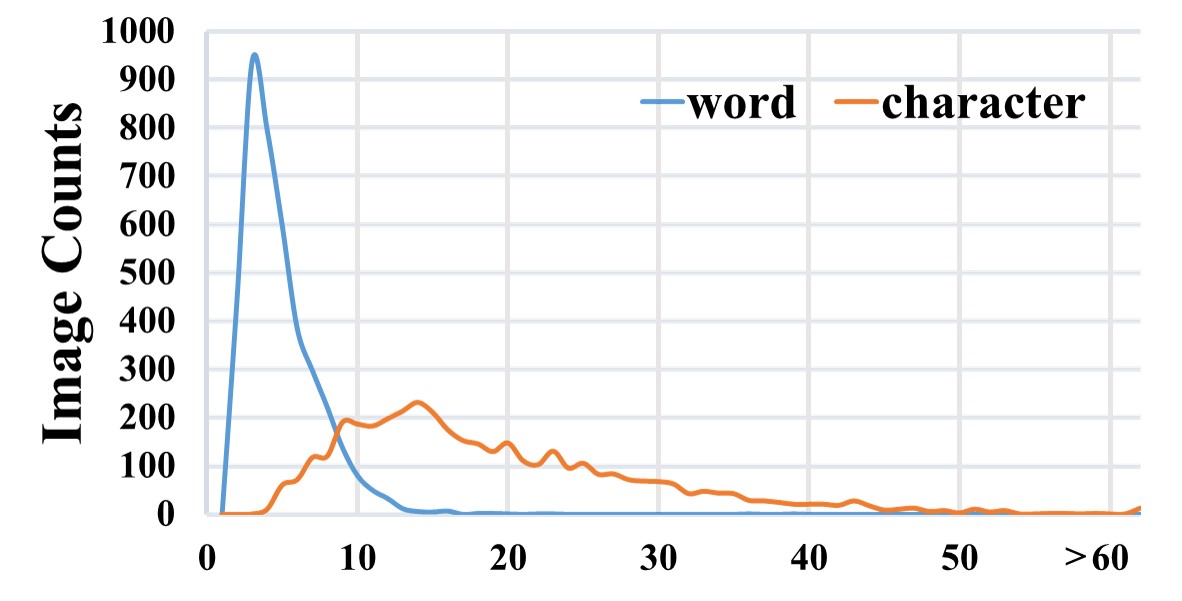}
        \vspace{-0.6cm}
        \caption{Number of objects}
        \label{fig:ds_stat1}
    \end{subfigure}
    \hspace{0.5mm}
    \begin{subfigure}{.23\textwidth}
        \centering
        \includegraphics[width=.99\linewidth]{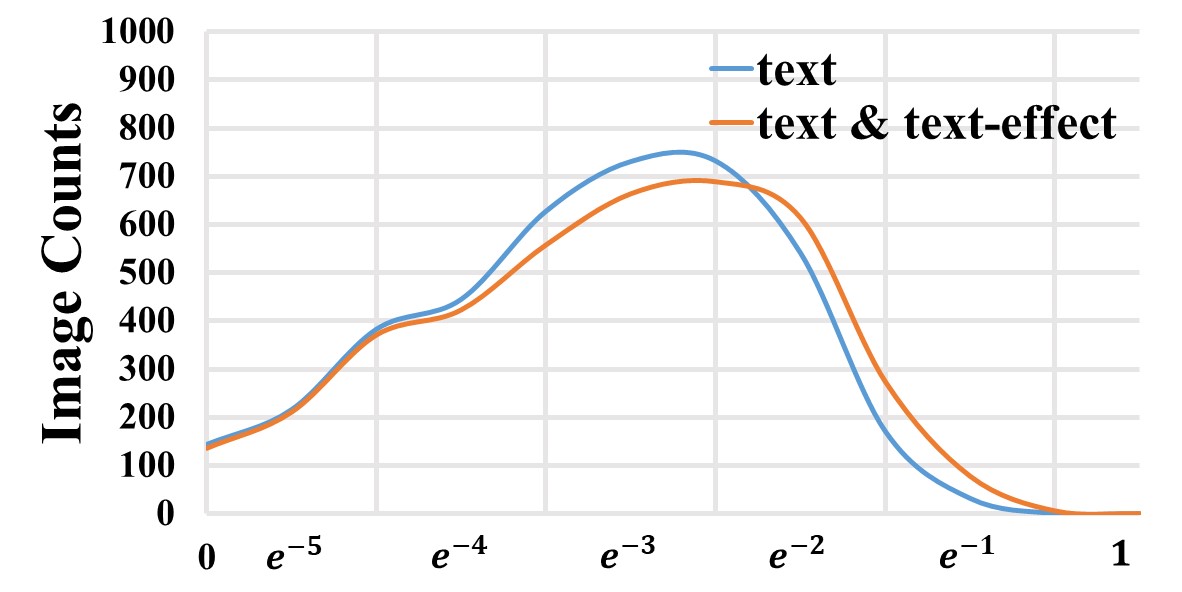}
        \vspace{-0.6cm}
        \caption{Text coverage ratio}
        \label{fig:ds_stat2}
    \end{subfigure}
    \begin{subfigure}{.49\textwidth}
        \centering
        \includegraphics[width=.99\linewidth]{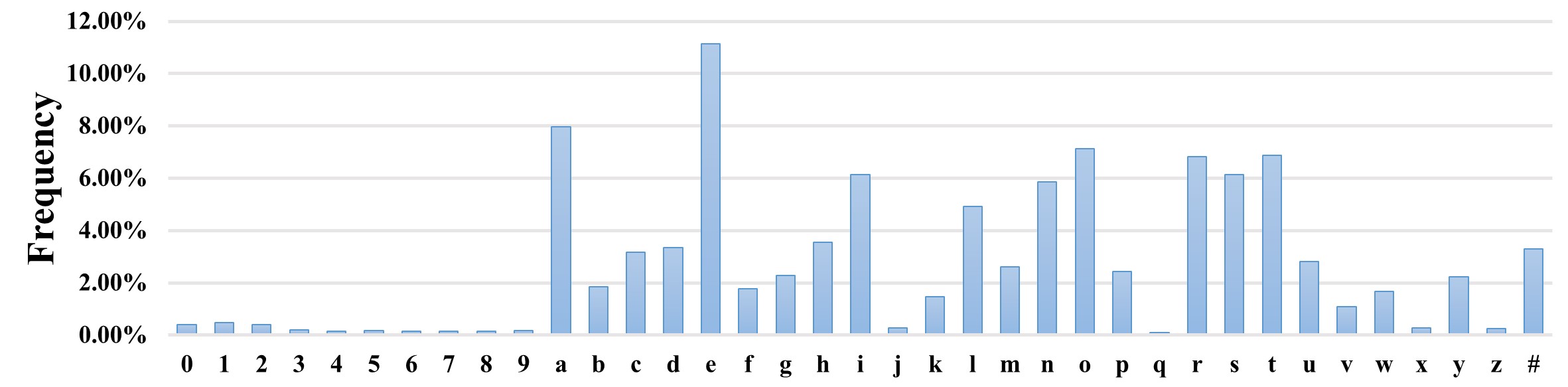}
        \vspace{-0.1cm}
        \caption{Letter frequency}
        \label{fig:ds_stat3}
    \end{subfigure}
\vspace{-0.2cm}
\caption{Statistics of \dsname{}. (a) Number of images with different numbers of words and characters. (b) Text coverage ratio against the image. (c) Character frequency of the whole dataset.}
\label{fig:ds_stat}
\end{figure}

\subsection{Statistical Analysis}

Statistical comparison between \dsname{} and four representative text segmentation datasets is listed in Table~\ref{table:ds_compare}, \ie, ICDAR13 FST~\cite{icdar13}, MLT\_S~\cite{mlts}, COCO\_TS~\cite{cocots}, and Total-Text~\cite{totaltext}.
In general, \dsname{} has more diverse text types and all types of annotations. Another dataset that provides character-level annotations is ICDAR13 FST, but its size is far smaller than other datasets. COCO\_TS and MLT\_S are relatively large, but they lack character-level annotations and mainly focus on scene text. The Total-Text was proposed \xxsub{recently }{}with similar scope to those existing datasets. 

\begin{figure*}[t!]
\centering
\begin{subfigure}{.19\textwidth}
  \centering
  \includegraphics[width=.98\linewidth]{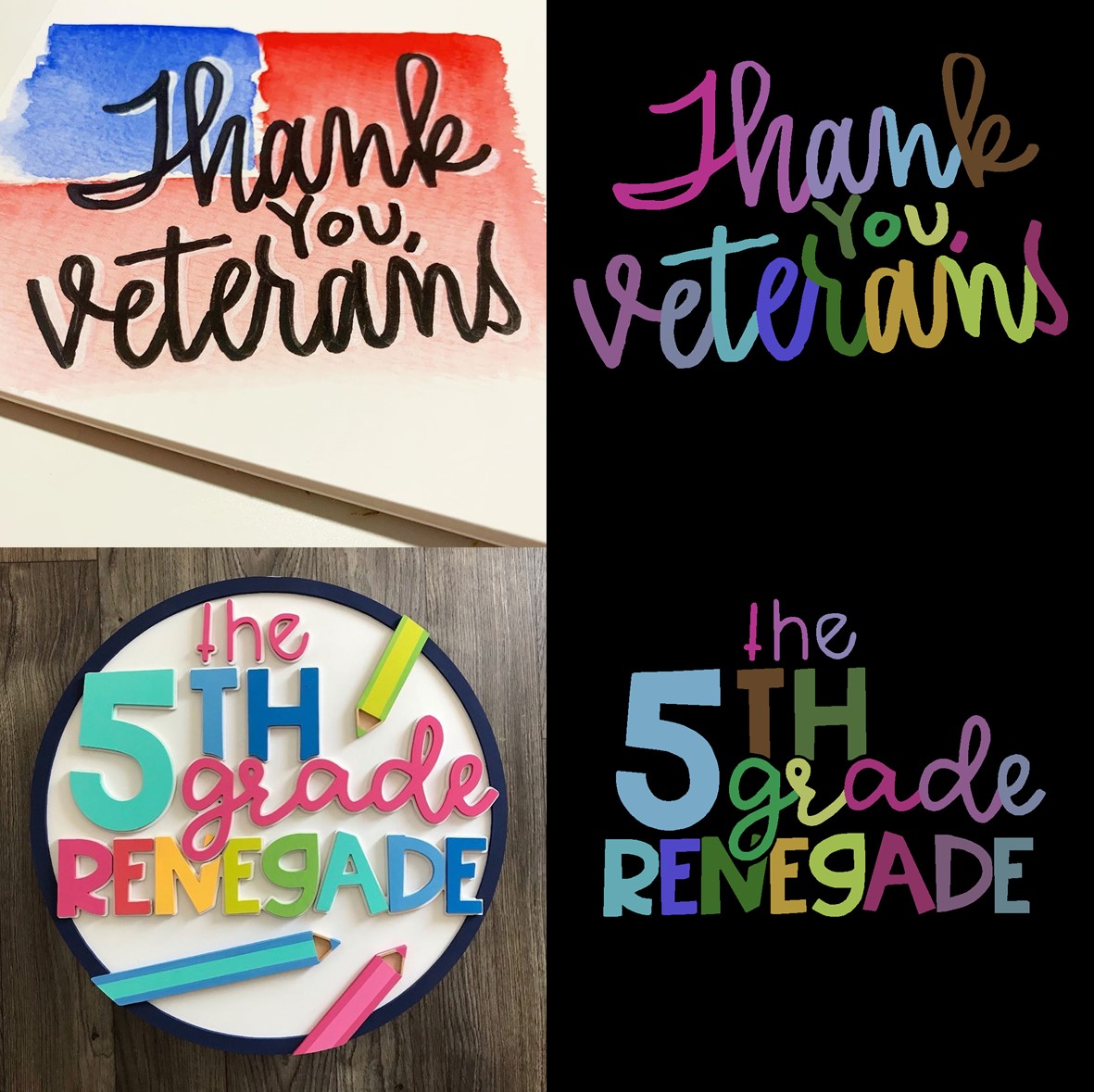}  
  \caption{\dsname{} (Ours)}
\end{subfigure}
\begin{subfigure}{.19\textwidth}
  \centering
  \includegraphics[width=.98\linewidth]{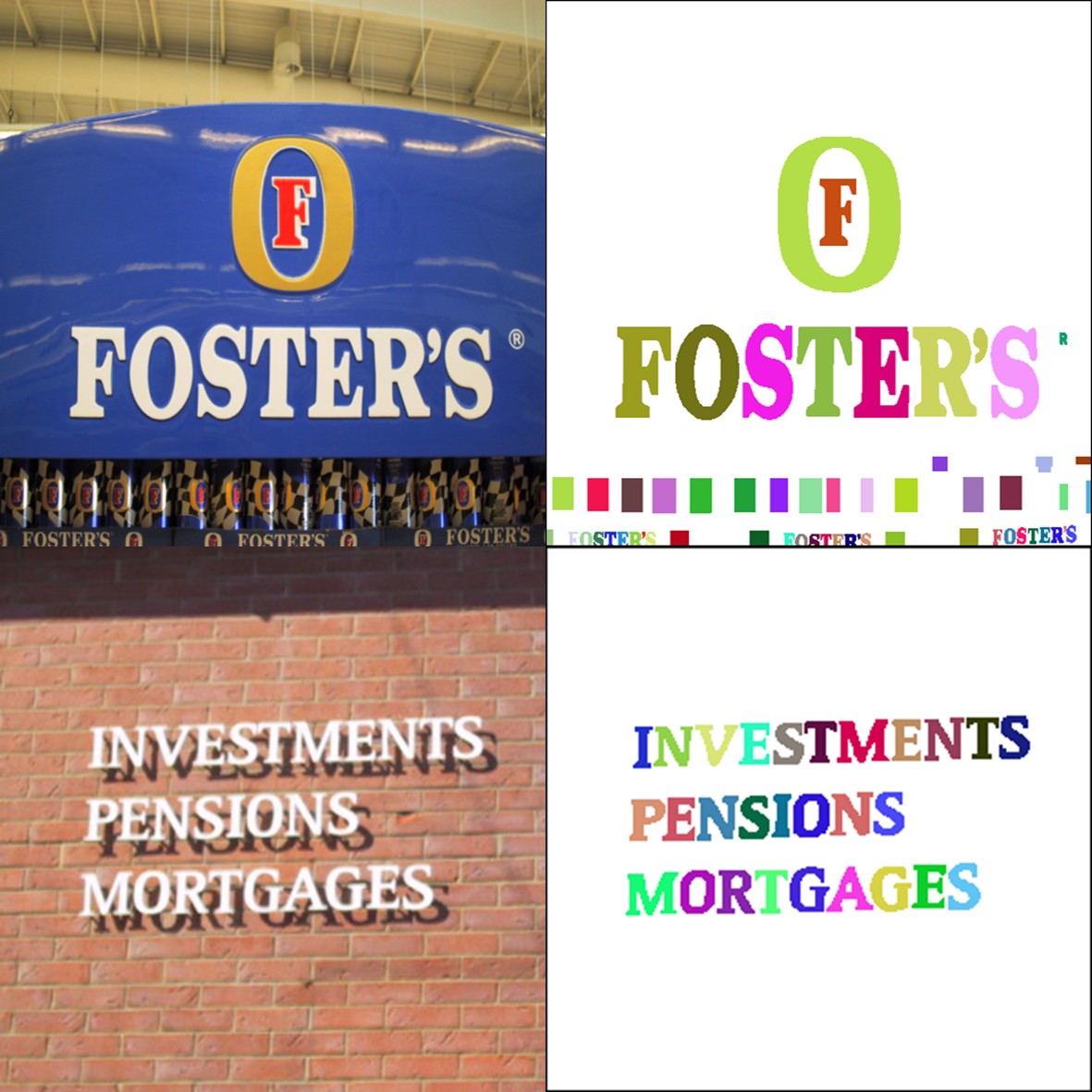}  
  \caption{ICDAR13 FST}
\end{subfigure}
\begin{subfigure}{.19\textwidth}
  \centering
  \includegraphics[width=.98\linewidth]{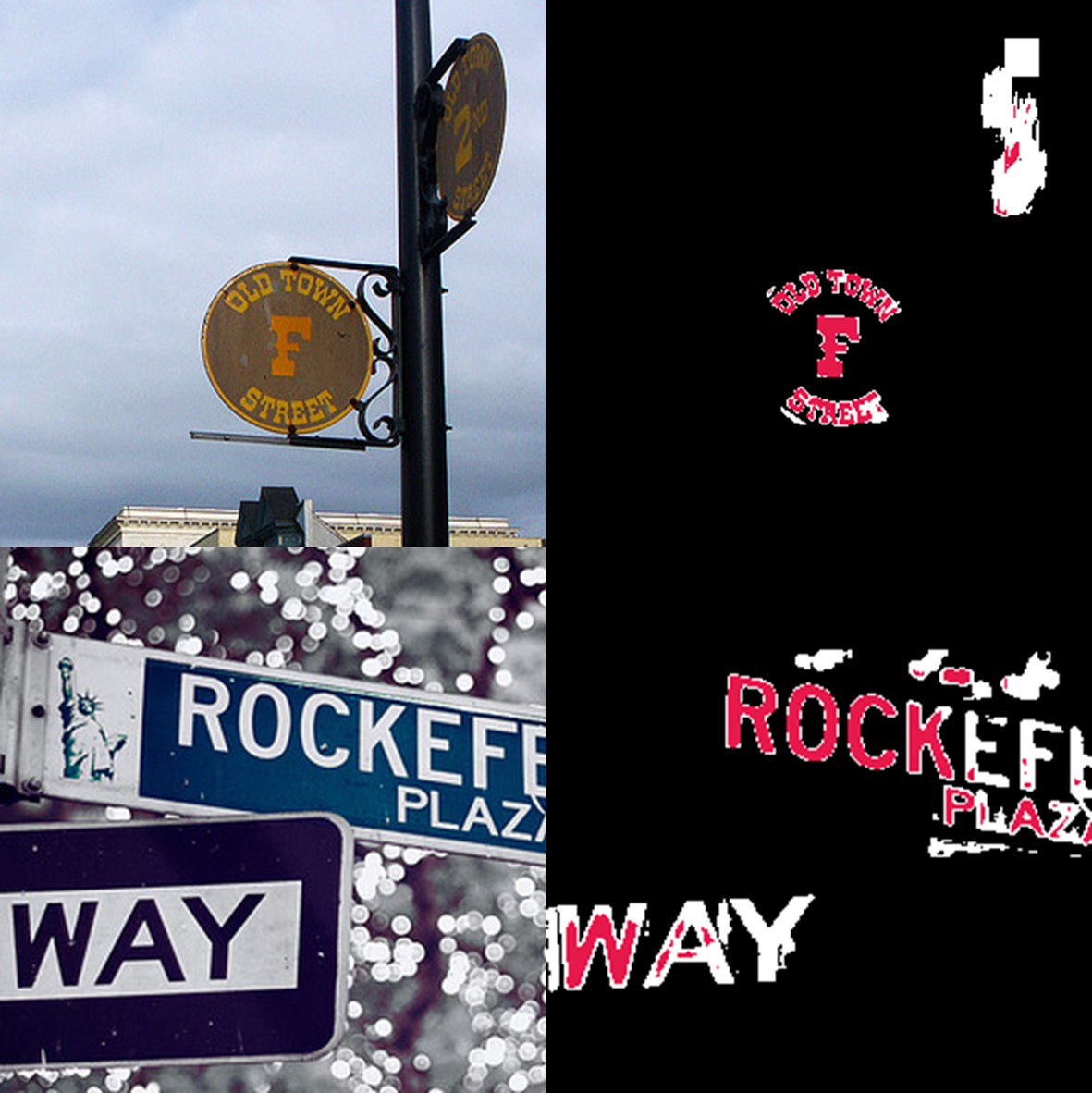}
  \caption{COCO\_TS}
\end{subfigure}
\begin{subfigure}{.19\textwidth}
  \centering
  \includegraphics[width=.98\linewidth]{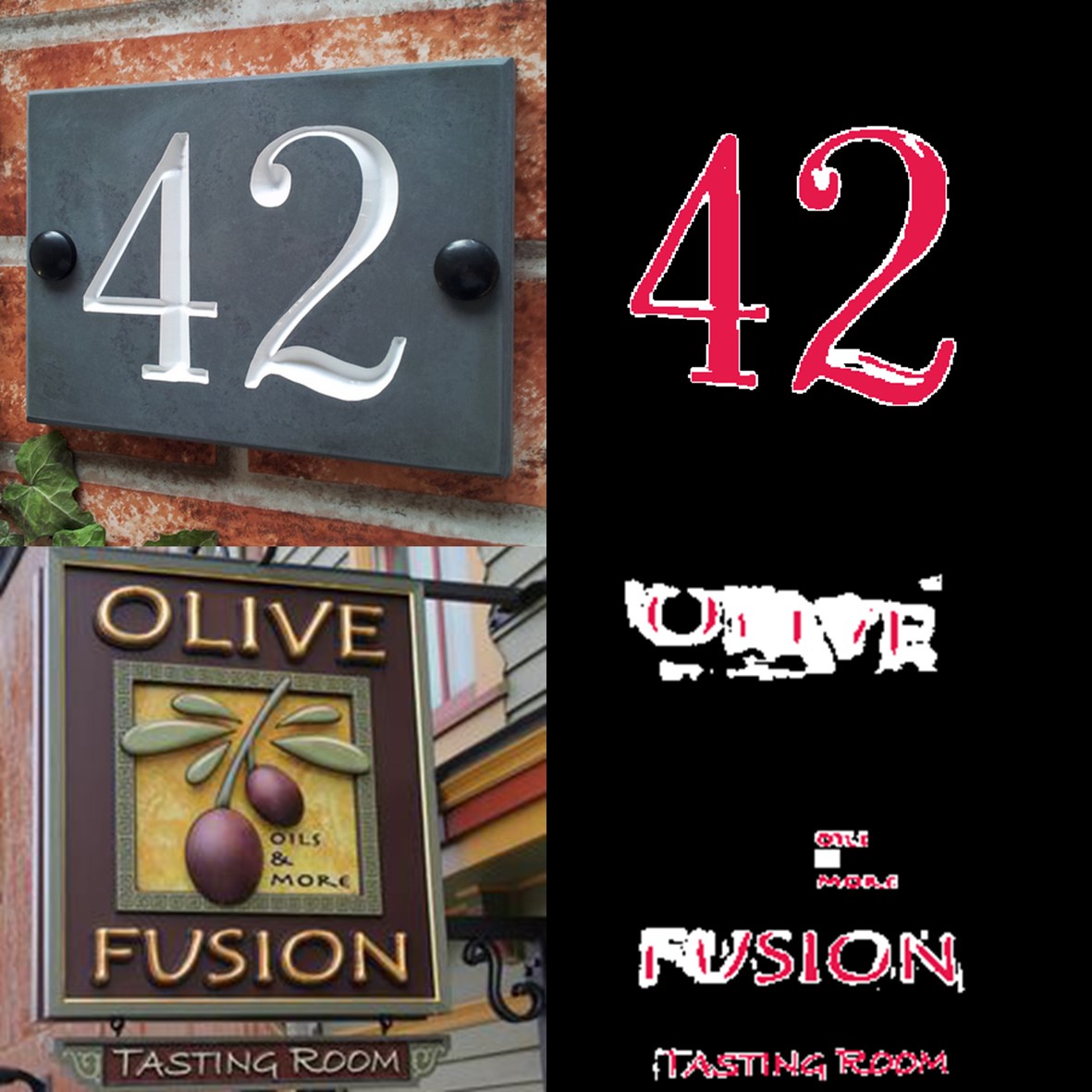} 
  \caption{MLT\_S}
\end{subfigure}
\begin{subfigure}{.19\textwidth}
  \centering
  \includegraphics[width=.98\linewidth]{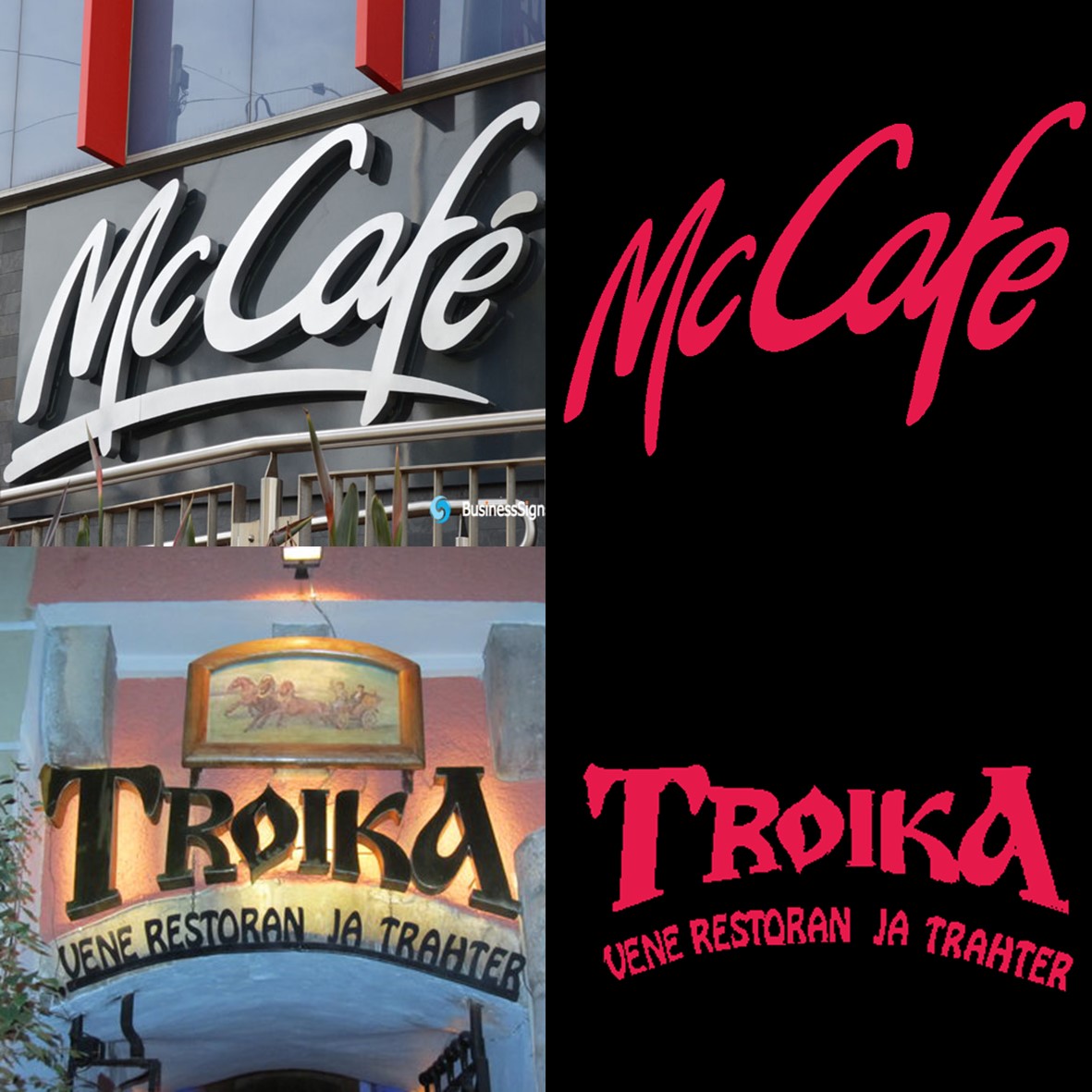}  
  \caption{Total-Text}
\end{subfigure}
\caption{Comparison of annotations from multiple text segmentation datasets. The proposed \dsname{} and ICDAR13 FST~\cite{icdar13} provide character-level annotations (color-coded characters). COCO\_TS~\cite{cocots}, MLT\_S~\cite{mlts}, and Total-Text~\cite{totaltext} only provide word-level annotations, where masks in red and white denote text regions and ignored regions, respectively. }
\label{fig:anno_compare}
\end{figure*}

\begin{table*}[t!]
\centering
\resizebox{0.95\textwidth}{!}{
    \begin{tabular}{
            L{5cm}
            |C{1.1cm}C{1.1cm}
            |C{1.1cm}C{1.1cm}
            |C{1.1cm}C{1.1cm}
            |C{1.1cm}C{1.1cm}
            |C{1.1cm}C{1.1cm}}
        \toprule
            & \multicolumn{2}{c|}{\dsname{} (Ours)}
            & \multicolumn{2}{c|}{ICDAR13 FST}
            & \multicolumn{2}{c|}{COCO\_TS}
            & \multicolumn{2}{c|}{MLT\_S}
            & \multicolumn{2}{c}{Total-Text}\\
            Method 
            & fgIoU & F-score
            & fgIoU & F-score
            & fgIoU & F-score
            & fgIoU & F-score
            & fgIoU & F-score\\ 
        \midrule
        PSPNet$^\dagger$~\cite{pspnet, cocots}
        & -- & --
        & -- & 0.797 
        & -- & --
        & -- & --  
        & -- & 0.740 \\
        SMANet$^\dagger$~\cite{mlts}
        & -- & --
        & -- & 0.785 
        & -- & --
        & -- & --  
        & -- & 0.770 \\
        % Tang and Wu^{\ddagger}~\cite{cnn_texseg_ccnn}
        % & -- & --
        % & -- & \textbf{\textit{0.895}} 
        % & -- & --
        % & -- & --  
        % & -- & -- \\
        DeeplabV3+~\cite{deeplabv3+}
        & 84.07 & 0.914 % good
        & 69.27 & 0.802 % good
        & 72.07 & 0.641 % good
        & 84.63 & 0.837 % good 
        & 74.44 & 0.824 % good
        \\
        HRNetV2-W48~\cite{hrnet}
        & 85.03 & 0.914 % good
        & 70.98 & 0.822 % good
        & 68.93 & 0.629 % good 
        & 83.26 & 0.836 % good
        & 75.29 & 0.825 % good 
        \\
        HRNetV2-W48 + OCR~\cite{ocr_seg}
        & 85.98 & 0.918 % good
        & 72.45 & 0.830 % good 
        & 69.54 & 0.627 % good  
        & 83.49 & 0.838 % good
        & 76.23 & 0.832 % good
        \\
        \midrule
        Ours: \modelname{} + DeeplabV3+
        & 86.06 & 0.921 % good
        & 72.16 & 0.835  % good
        & \textbf{73.98} & \textbf{0.722} % good 
        & \textbf{86.31} & 0.860 % good 
        & 76.53 & 0.844 % good
        \\
        Ours: \modelname{} + HRNetV2-W48
        & \textbf{86.84} & \textbf{0.924} % good
        & \textbf{73.38} & \textbf{0.850} % good
        & 72.39 & 0.720 % good 
        & 86.09 & \textbf{0.865} % good 
        & \textbf{78.47} & \textbf{0.848} % good 
        \\
        \bottomrule
    \end{tabular}
}
\vspace{0.1cm}
\resizebox{0.95\textwidth}{!}{
    \begin{tabular}{L{0.2cm}L{24cm}}
        $^\dagger$ 
        & In ~\cite{cocots, mlts}, the author augmented the original training dataset with SynthText~\cite{syntex} in both ICDAR13 FST and Total-Text experiments.
        \\
        % $\ddagger$ 
        % & In ~\cite{cnn_texseg_ccnn}, the author augmented their training with an non-public synthetic dataset and evaluated in ICDAR13 FST.
    \end{tabular}
}

\caption{
    % This table compares the performances of our models with the performances of other semantic segmentation models on 4 other datasets. 
    Performance comparison between \modelname{} and other models on \dsname{} and other representative text segmentation datasets. The bold numbers indicate the best results.
}
\label{table:perf_alldata}
\vspace{-0.3cm}
\end{table*}

The 4,024 images in \dsname{} are split into training, validation, and testing sets with 2,646, 340, and 1,038 images, respectively. In \dsname{} and all its splits, the ratio between the number of scene text and design text is roughly 1:1. Figure~\ref{fig:ds_stat1} counts the number of images with different numbers of words and characters, where 12-16 characters and 2-4 words per image is the majority. Figure~\ref{fig:ds_stat2} shows the distribution of the text coverage ratio, where the blue line is set up for word masks \xxsub{while}{and} the orange line is for word-effect masks. The rightward shifting from blue to orange indicates the coverage increment due to the word-effect. Finally, Figure~\ref{fig:ds_stat3} displays the character frequency in \dsname{}, which roughly aligns with that of English corpus. 
\subsection{Qualitative Comparison}

Figure~\ref{fig:anno_compare} shows qualitative comparison between \dsname{}, ICDAR13 FST, COCO\_TS, MLT\_S and Total-Text. ICDAR13 FST has many box-shape character masks (considered as ignored characters), which is not a common case in the proposed \dsname{}. The other datasets, \ie, COCO\_TS, MLT\_S, and Total-Text, have only word masks. Note that COCO\_TS and MLT\_S introduce a large number of ignored areas, especially along text boundaries, which would hinder models from precisely predicting text boundaries. \xxsub{Those boundary-ignored annotations may be caused by automatic labeling.}{Those boundary-ignored annotations are caused by automatic labeling using weekly supervised models.} Similar to \dsname{}, Total-Text is labeled manually, but it is of a much smaller size than ours and lacks annotations of characters and text effects.

\vspace{-0.2cm}
\section{Experimental Evaluation}
\label{sec:experiment}

To demonstrate the effectiveness of the proposed \modelname{}, it will be compared to the state-of-the-art methods DeeplabV3+~\cite{deeplabv3+} and HRNet-W48~\cite{hrnet} on five datasets, \ie, ICDAR13 FST~\cite{icdar13}, COCO\_TS~\cite{cocots}, MLT\_S~\cite{mlts}, Total-Text~\cite{totaltext}, and the proposed \dsname{}. 
%\bp{We can delete the rest of this paragraph if we need space.} \xx{deleted} \xxsub{The experiment setup and evaluation metrics are described in Section~\ref{subsec:setup}, and then ablation studies are conducted to validate the proposed losses and network design in Section~\ref{subsec:ablation}. Quantitative comparison to other algorithms is shown in Section~\ref{subsec:cmp}. Finally, two representative downstream applications are raised in Section~\ref{subsec:applications} to illustrate the advantages of leveraging text segmentation in these works.}{}

\subsection{Experiment Setup}
\label{subsec:setup}

Each model in comparison will be re-trained on each of the aforementioned text segmentation datasets. The models are initialized by ImageNet pretrains and then trained on 4 GPUs in parallel using SGD with weight decay of $5e^{-4}$ for 20,500 iterations. The first 500 iterations are linear warm-ups~\cite{linear_warmup}, and the rest iterations use poly decayed learning rates starting from 0.01~\cite{deeplabv3+}. Note that 5,500 iterations are performed on ICDAR13 FST due to its small size as shown in Table~\ref{table:ds_compare}. For \dsname{}, our model train and evaluate using word masks as foreground instead of the word-effect masks.

The glyph discriminator in \modelname{} adopts a ResNet50 classifier~\cite{resnet}, which is trained on character patches from \dsname{} training and validation sets. It achieves the classification accuracy of 93.38\% on the \dsname{} testing set. Since only the proposed \dsname{} and ICDAR13 FST provide character bounding boxes, the glyph discriminator is only applied on these two datasets and disabled on COCO\_TS, MLT\_S, and Total-Text.

To align with modern segmentation tasks, we use foreground Intersection-over-Union (fgIoU) as our major metric. In addition, \xxsub{we report F-score on foreground pixels.}{the typical F-score measurement on foreground pixels are provided in the same fashion as~\cite{fscore_txt, icdar13}. } The foreground here indicates the text region in both prediction and ground truth.

\subsection{Model Performance}
\label{subsec:cmp}

This section compares \modelname{} to other text and semantic segmentation methods. To demonstrate the effectiveness of \modelname{}, the comparison is conducted on five datasets including our \dsname{}. As previously claimed, we adopt DeeplabV3+~\cite{deeplabv3+} and HRNetV2-W48~\cite{hrnet} as our backbone and baseline. We also compares with the SOTA semantic segmentation model: HRNetV2-W48 + Object-Contextual Representations (OCR)~\cite{ocr_seg}. The PSPNet and SMANet results are from~\cite{cocots, mlts} in which their models were trained on ICDAR13 FST and Total-Text augmented with SynthText~\cite{syntex}. Tables~\ref{table:perf_alldata} shows the overall results. As the table shows, our proposed \modelname{} outperforms other methods on all datasets.

% \input{table/perf_otherdata}

%\xxsub{In general, \modelname{} outperforms other methods on all datasets although the performance is slightly affected by the backbone. More specifically, \modelname{} + DeeplabV3+ performs better on COCO\_TS and ICDAR17 MLT\_S, while \modelname{} + HRNet-W48 achieves higher scores on \dsname{}, ICDAR13 FST, and Total-Text. The general improvement from \modelname{} is around 2.3\% as compared to the backbone. The maximum improvement is 3.18\% on Total-Text, and the minimum improvement is 1.68\% on ICDAR17 MLT\_S.}{}

\subsection{Ablation Studies}
\label{subsec:ablation}

This section performs ablation studies on the key pooling and attention (the yellow block in Figure~\ref{fig:network}), trimap loss, and glyph discriminator in the proposed \modelname{}. In this experiment, DeeplabV3+ is adopted as the backbone, and the models are trained and evaluated on \dsname{}. Starting from the base version of \modelname{}, the key pooling and attention (Att.), trimap loss ($\mathcal{L}_{tri}$), and glyph discriminator ($\mathcal{L}_{dis}$) are added incrementally as shown in Table~\ref{table:ablation}, where the fgIoU and F-score are reported, presenting a consistently increasing trend.  
%\xxsub{The \modelname{} (base) removes the attention mechanism by skipping $x_{sem}$ to the concat layer, \ie, directly concatenating $x_{sem}$ to the input image $x$ and feature map $x_f$ and feeding them to the conv5x5 layer. Note that the number of parameters in \modelname{} (base) is the same as that in \modelname{}.}{}
The final \modelname{} achieves the best performance, around 2\% increase in fgIoU as compared to DeeplabV3+.

\begin{table}[h!]
\centering
\resizebox{0.45\textwidth}{!}{
    \begin{tabular}{R{2.4cm}|C{0.8cm}C{0.8cm}C{0.8cm}|C{1cm}C{1.2cm}}
        \toprule
        Method & Att. & $\mathcal{L}_{tri}$ & $\mathcal{L}_{dis}$ & fgIoU & F-score \\ 
        \midrule
        DeeplabV3+
            & & & 
            & 84.07 & 0.914 % good
            \\
        \modelname{} (base)  
            & & & 
            & 84.86 & 0.917 % good
            \\
        \modelname{}  
            & $\checkmark$ & & 
            & 85.36 & 0.919 % good
            \\
        \modelname{}  
            & $\checkmark$ & $\checkmark$ & 
            & 85.55 & 0.921 % good
            \\
        \midrule
        \modelname{} (final) 
            & $\checkmark$ & $\checkmark$ & $\checkmark$ 
            & \textbf{86.06} & \textbf{0.921} % good 
            \\
        \bottomrule
    \end{tabular}
}
\vspace{0.2cm}
\caption{
    Ablation studies of \modelname{} on \dsname{}. All models are training on \dsname{} train and validation sets, and  all \modelname{} use DeeplabV3+ as backbone. The column ``Attn.'' represents whether attention layers are included. Similarly, columns ``$\mathcal{L}_{tri}$'' and ``$\mathcal{L}_{dis}$'' indicate whether the trimap loss and glyph discriminator are used. 
}
\label{table:ablation}
\end{table}

%\xxsub{From Table~\ref{table:ablation}, the final \modelname{} achieves the best performance, around 2\% increase in fgIoU as compared to DeeplabV3+. An interesting observation is that \modelname{} (base) and \modelname{} (final) have exactly the same number of parameters, while the improvement from \modelname{} (base) to \modelname{} (final) is more than that from DeeplabV3+ to \modelname{} (base). In other words, the proposed non-parametric attention mechanism and losses contributes more than expanding convolutional layers to the backbone.}{}

\begin{figure}[t!]
    \centering
    \includegraphics[width=0.45\textwidth]{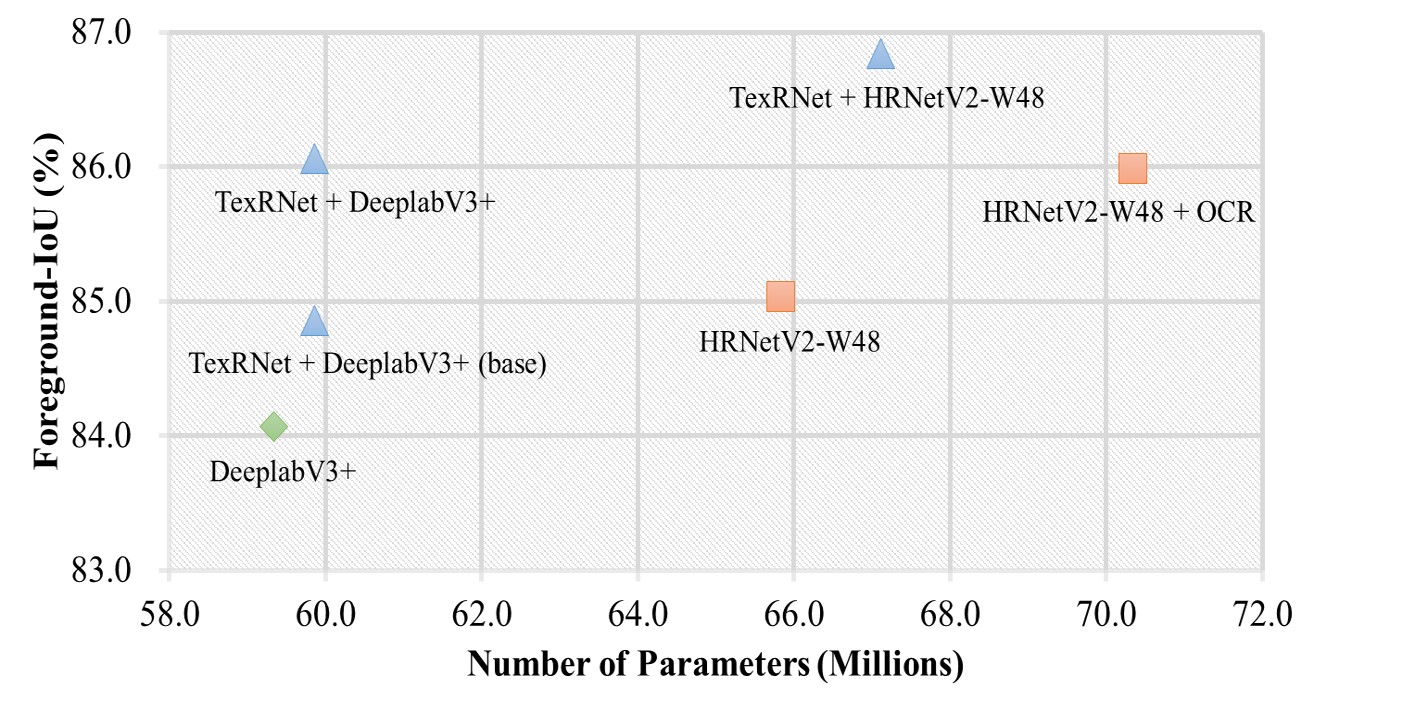}
    \vspace{-0.2cm}
    \caption{
    %Our model \modelname{} outperforms DeeplabV3+, HRNetV2-W48 and HRNetV2-W48+OCR on \dsname{} in terms of fgIoU using less parameters.
    Comparison of different methods in the number of parameter vs. text segmentation performance in fgIoU.
    }
    \vspace{-0.1cm}
\label{fig:paranum}
\end{figure}

An interesting observation is that \modelname{} (final) have exactly the same number of parameters as \modelname{} (base), but the part between them contributes the most improvement. To further investigate whether the performance increase comes from parameter increase, we compared \modelname{} with HRNetV2-W48+OCR and other models in Figure~\ref{fig:paranum}. We discover that \modelname{} achieves higher accuracy with less parameters as compared to HRNetV2-W48+OCR, demonstrating the effectiveness of our design in \modelname{}. 

\subsection{Downstream applications}
\label{subsec:applications}

This section gives prospects of \modelname{} and \dsname{} dataset, especially in driving downstream applications. 
%\xxsub{The following introduced two representative tasks, \ie, scene text removal and scene text style transfer, which would significantly benefit from text segmentation.}{}

\begin{figure}[h!]
    \centering
    \includegraphics[width=0.45\textwidth]{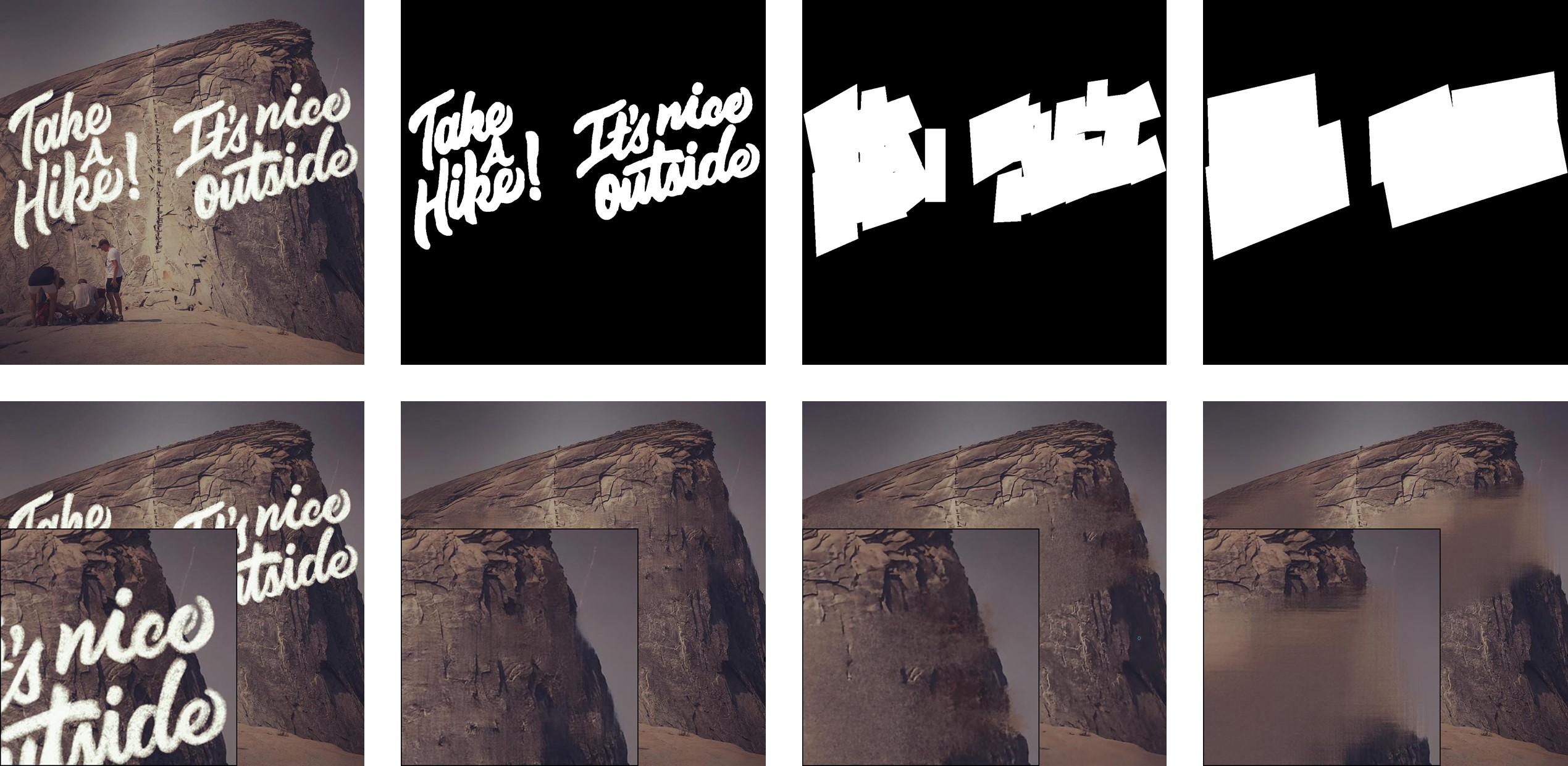}
    \caption{
    % This figure shows our text removal results on images in \dsname{}. The left most column shows the original images. From the second to the last column are text removal visualizations using predicted text masks from our \modelname{}, character ground truth bounding polygons and word ground truth bounding polygons. We also provide zoom-in windows at bottom-left for quality inspections.
    Examples of text removal with different types of text masks. From left to right, the top row shows the input image, predicted text mask from our \modelname{}, character bounding polygons, and word bounding polygons from ground truth. The second row are text removing results using corresponding text masks on the same column. 
    %\zz{it is still better to show text masks in each case.}
    }
\label{fig:text_removal}
\vspace{-0.2cm}
\end{figure}

\textbf{Text Removal} is a practical problem in photo and video editing, \xxsub{}{and it is also an application with high industrial demand. For example, media service providers frequently need to erase brands from their videos to avoid legal issues.} Since this task is a hole filling problem, Deep Image Prior~\cite{deep_image_prior} is employed, and different types of text masks are provided to compare the performance of text removal. Typically, word or character bounding boxes are standard text masks because they can be easily obtained from existing text detection methods. By contrast, the proposed \modelname{} provides much more accurate text masks. Figure~\ref{fig:text_removal} compares the results using these three types of text masks, \ie, text segmentation mask, character bounding polygon, and word bounding polygon. 
%Note that the character and word bounding boxes are directly obtained from the ground truth. 
Obviously, finer mask results in better performance because of more retention on backgrounds, and \modelname{} provides the finest text mask than the others. For more examples, please refer to the supplementary.

\textbf{Text Style Transfer} is another popular task for both research and industry. Mostly, text style transfer relies on accurate text masks. In this experiment, we use Shape-Matching GAN~\cite{shapemgan} as our downstream method, which requires text masks as an input. In their paper, all demo images are generated using ground truth text masks, which may be impractical in real-world applications. Therefore, we extend \modelname{} with Shape-Matching GAN to achieve scene text style transfer on an arbitrary text image. A few examples are visualized in Figure~\ref{fig:style_transfer}, and more examples can be found in the supplementary.

\begin{figure}[t!]
    \centering
    \includegraphics[width=0.45\textwidth]{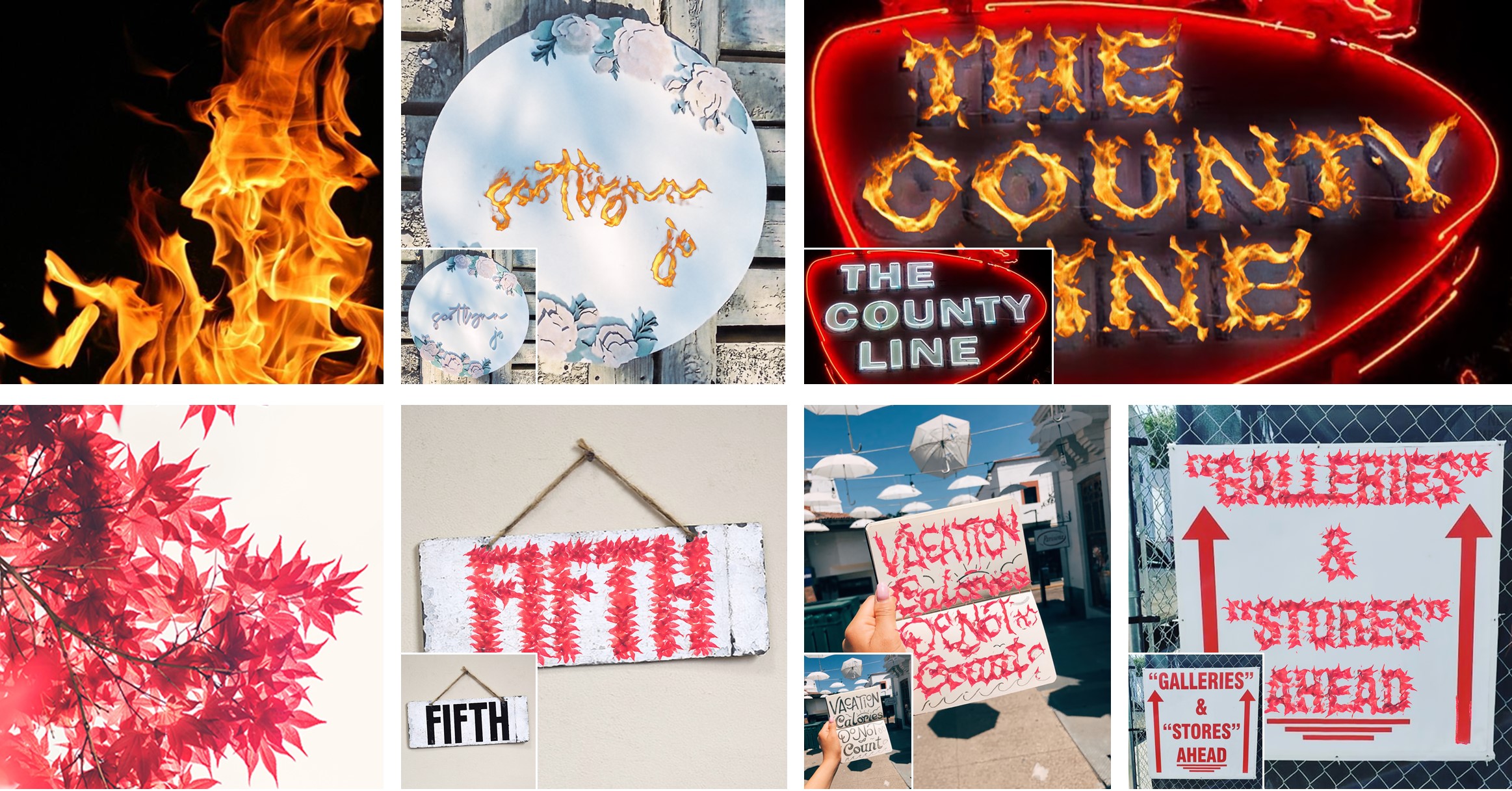}
    \caption{
    % This figure shows our style transfer results on text images. 
    Examples of text style transfer with styles of fire and maple on the first column. The rest are results with their original images attached to the bottom-left corner.
    % \zz{we can show fewer examples, but original and style images are necessary}
    }
\label{fig:style_transfer}
\vspace{-0.2cm}
\end{figure}

\vspace{-0.2cm}
\section{Conclusion}
We introduce a novel text segmentation dataset \dsname{}, which consists of 4,024 scene text and design text images with comprehensive annotations including word- and character-wise bounding polygons, masks and transcriptions. Moreover, we purpose a new and effective text segmentation method \modelname{}, and we demonstrate that our model outperforms state-of-the-art semantic segmentation models on \dsname{} and another four datasets. To support our idea that text segmentation has great potential in industry, we introduce two downstream applications, \ie, text removal and text style transfer, to show promising results using text segmentation masks from \modelname{}. In conclusion, text segmentation is a critical task. We hope that our new dataset and method would become the corner-stone for future text segmentation research. 

% \bp{The text segmentation module name above is missing.} \xx{added}

% \input{figure/ds_stat}

{\small
\bibliographystyle{ieee_fullname}
\bibliography{egbib}
}

% \bibliographystyle{ieee_fullname}
% \bibliography{egbib}
% \clearpage

\appendix
\section*{Appendix}
\renewcommand{\thesubsection}{\Alph{subsection}}

\section{Cosine Similarity vs. Accuracy}

\begin{figure}[h!]
\centering
\includegraphics[width=.49\textwidth]{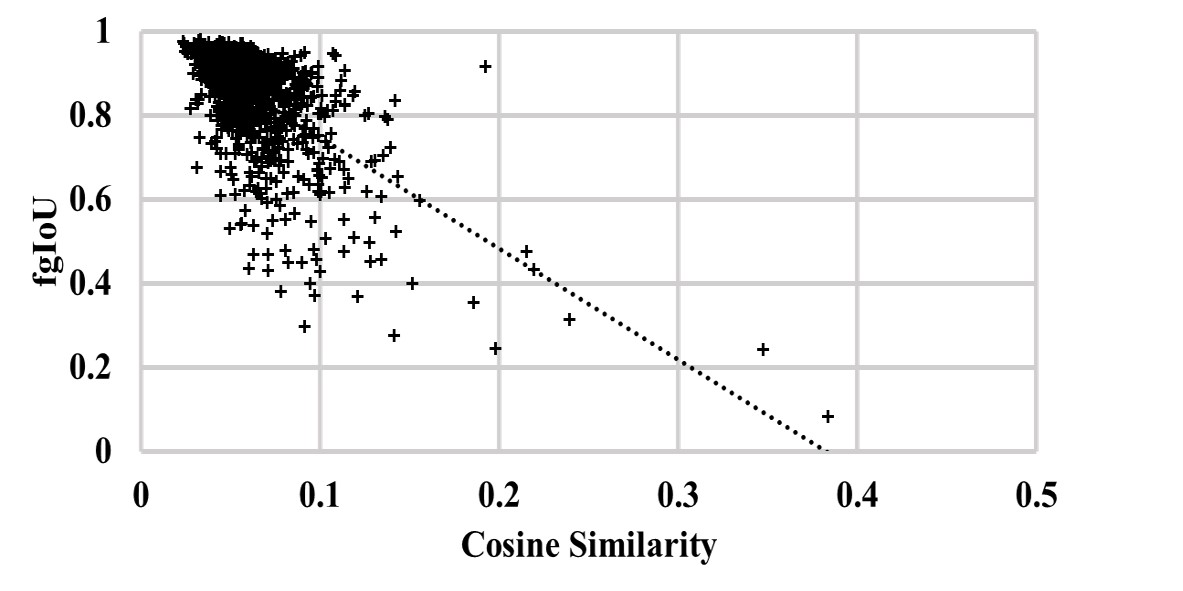}
\caption{The relation between cosine similarity and fgIoU from predicted masks. In our case, the cosine similarity is computed from the vectorized text prediction and vectorized background prediction.}
\label{fig:cossim_vs_acc}
\end{figure}

Recall that in our paper, we claim that the cosine similarity of the predicted mask is inversely correlated with its accuracy. To provide solid evidence on our claim, we produce the following experiment in which we compute the cosine-similarity, \ie CosSim($x'_{sem}$), and the fgIoU score on each image and plot their relation in Figure~\ref{fig:cossim_vs_acc}. According to our plot, there is a clear downward trend between Cosine Similarity and fgIoU, from which our claim can be verified.

\section{\dsname{} Domain Studies}

To further show that our \dsname{} is a strong complementary towards many prior datasets, we performed domain studies in the same fashion as~\cite{mlts}. The goals of our experiments are two-fold, a) to compare fairly with SMANet~\cite{mlts} on ICDAR13 FST~\cite{icdar13} and Total-Text~\cite{cocots} under the same dataset setting, and b) to show the performance boost by including our dataset \dsname{} in the training process. The experiments were carried out using our proposed \modelname{} with nearly the same experiment settings as explained in Session 5.1 of the main paper. The only differences are: we disabled the discriminator loss (\ie $\mathcal{L}_{dis}$) and we used 20,500 iterations in the ICDAR13 FST experiments. Such changes were due to the fact that no character bounding polygons were provided in COCO\_TS~\cite{cocots}, MLT\_S~\cite{mlts}, and more images could be used to avoid overfits. As Table~\ref{table:ds_icdar13} and Table~\ref{table:ds_tt} shows, our \modelname{} reached F-score 0.866 on ICDAR13 FST and 0.844 on Total-Text, exceeding SMANet using the same dataset combination in training. Meanwhile, we demonstrated an extra 3.20\% and 2.93\% increase in fgIoU when included our \dsname{} in training on ICDAR13 FST and Total-Text, correspondingly.

\begin{table*}[h!]
\centering
\resizebox{0.8\textwidth}{!}{
    \begin{tabular}{
            L{2.5cm}L{8.5cm}C{1.1cm}C{1.1cm}}
        \toprule
            Method & Train Dataset & fgIoU & F-score\\ 
        \midrule
            \multirow{3}{0pt}{SMANet~\cite{mlts}} & ICDAR13 FST & 
                - & 0.713 \\
            & ICDAR13 FST + Synth~\cite{syntex} & 
                - & 0.785 \\
            & ICDAR13 FST + COCO\_TS + MLT\_S & 
                - & 0.858 \\
        \midrule
            \multirow{4}{0pt}{\modelname{} (Ours)} & ICDAR13 FST & 
                73.38 & 0.850 \\
            & ICDAR13 FST + COCO\_TS + MLT\_S & 
                76.68 & 0.866 \\
            & ICDAR13 FST + COCO\_TS + MLT\_S + \dsname{} (ours) & 
                78.65 & 0.871 \\
            & \textbf{ICDAR13 FST + \dsname{} (Ours)} & 
                \textbf{79.88} & \textbf{0.887} \\
        \bottomrule
    \end{tabular}
}
\caption{
    Domain studies on ICDAR13 FST in which models are training with different datasets and are evaluated on ICDAR13 FST test set. 
}
\label{table:ds_icdar13}
\end{table*}

\begin{table*}[h!]
\centering
\resizebox{0.8\textwidth}{!}{
    \begin{tabular}{
            L{2.5cm}L{8.5cm}C{1.1cm}C{1.1cm}}
        \toprule
            Method & Train Dataset & fgIoU & F-score\\ 
        \midrule
            \multirow{3}{0pt}{SMANet~\cite{mlts}} 
            & Total-Text & 
                - & 0.741 \\
            & Total-Text + Synth~\cite{syntex} & 
                - & 0.770 \\
            & Total-Text + COCO\_TS + MLT\_S & 
                - & 0.781 \\
        \midrule
            \multirow{4}{0pt}{\modelname{} (Ours)} 
            & Total-Text & 
                78.47 & 0.848 \\
            & Total-Text + COCO\_TS + MLT\_S & 
                77.40 & 0.844 \\
            & Total-Text + COCO\_TS + MLT\_S + \dsname{} (Ours) & 
                80.01 & \textbf{0.858} \\
            & \textbf{Total-Text + \dsname{} (Ours)} & 
                \textbf{80.33} & 0.856 \\
        \bottomrule
    \end{tabular}
}
\caption{
    Domain studies on Total-Text in which models are training with different datasets and are evaluated on Total-Text test set. 
}
\vspace{-0.2cm}
\label{table:ds_tt}
\end{table*}

\section{Visual Comparison on \modelname{}}

To help to understand the key structure of our \modelname{}, we extract activation maps from intermediate stages of \modelname{} to show how the low confidence text regions in the initial prediction are re-activated using our key pooling and attention module.

\vspace{0.2cm}

\begin{figure}[h!]
\centering
\includegraphics[width=.45\textwidth]{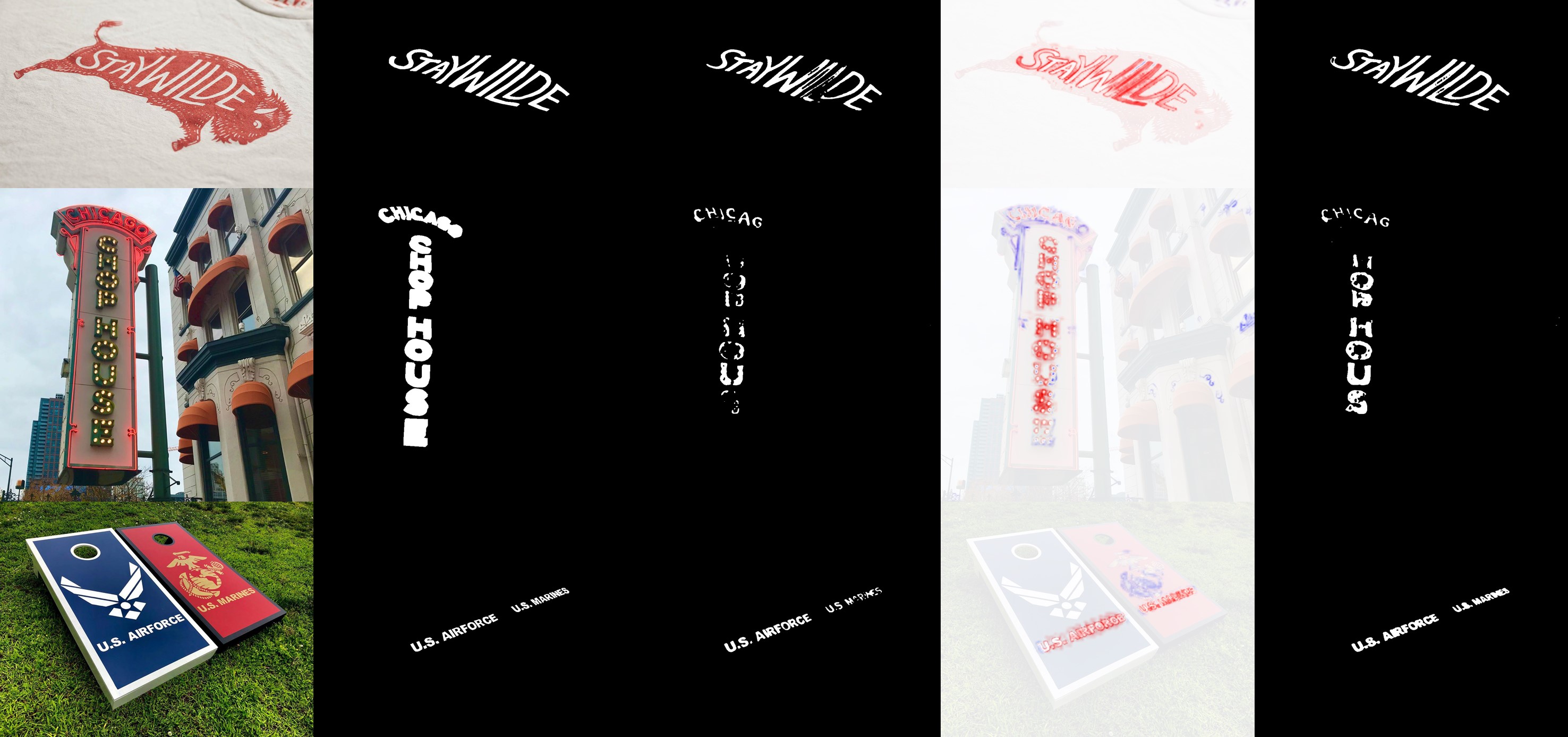}
\caption{Segmentation samples in which the key pooling and attention mechanism in \modelname{} helps re-activate low-confidence text regions and achieves better performance. From left to right are original images, ground truth labels, initial predictions, gradients of activation, and refined predictions. In the gradients of activation plot (\ie the $4^{th}$ column), red indicates positive score changes and blue indicates negative score changes.}
\label{fig:react}
\end{figure}

In particular, the $4^{th}$ column of Figure~\ref{fig:react} highlights the gradients of activation scores between the initial predictions (\ie $x'_{sem}$) and the activation maps prior to the concatenation and the refinement layers (\ie $x_{att}$).

\section{Visualization on Text Removal}

\begin{figure*}[t!]
\centering
\includegraphics[width=.91\textwidth]{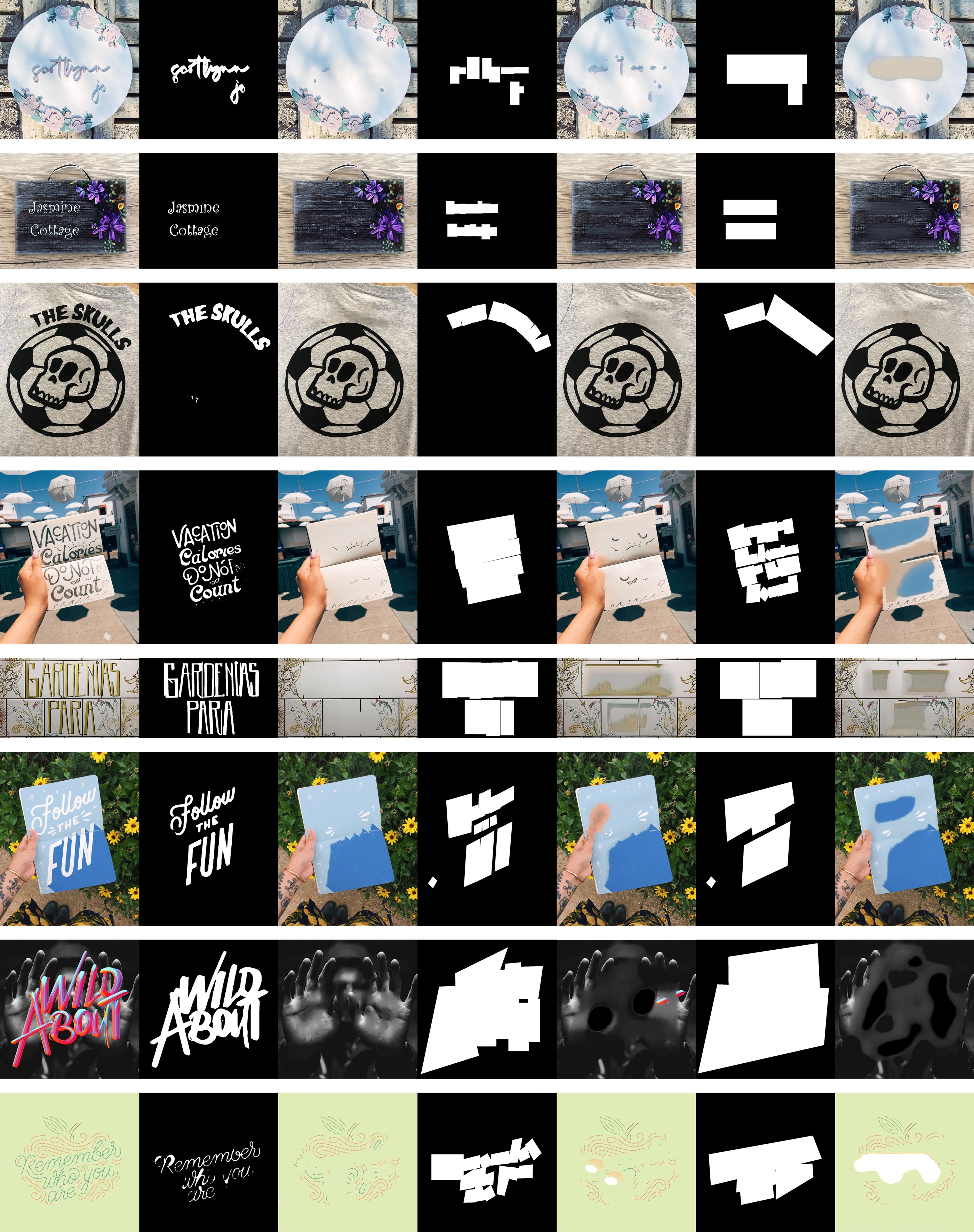}
\vspace{0.5cm}
\caption{Text removal visualization using predicted text masks from our \modelname{} and inpainting network Deep Image Prior~\cite{deep_image_prior}. For each sample, the left-to-right ordering of the plots are as such: original image; predicted mask; text removal using mask; ground truth character bounding polygon (char-bpoly); text removal using char-bpoly; ground truth word bounding polygon (word-bpoly); text removal using word-bpoly.}
\label{fig:tr_supp}
\end{figure*}

This session shows extra samples from our text removal experiment. Recall that we predicted text masks from our \modelname{} and used them as inputs for Deep Image Prior~\cite{deep_image_prior}. The \modelname{} was trained on \dsname{} train and validation sets, while all demo images were from \dsname{} test set. We also produced examples using ground truth bounding polygons as alternative inputs. As shown in Figure~\ref{fig:tr_supp}, text-free images generated using our predicted mask has the best performance. 

\section{Visualization on Text Style Transfer}

This session shows extra style transfer samples using predicted text masks from our \modelname{} and text style transfer network Shape-Matching GAN~\cite{shapemgan}. Same as text removal, our model was trained on \dsname{} train and validation sets, and predicted on the test set. For each sample, the original image, the predicted text mask, and the final result are shown from left to right. We show three styles in total, which is fire (Figure~\ref{fig:st_supp_fire}), maple (Figure~\ref{fig:st_supp_maple}) and water (Figure~\ref{fig:st_supp_maple}).

\begin{figure*}[t!]
\centering
\begin{subfigure}{.99\textwidth}
    \centering
    \includegraphics[width=.99\textwidth]{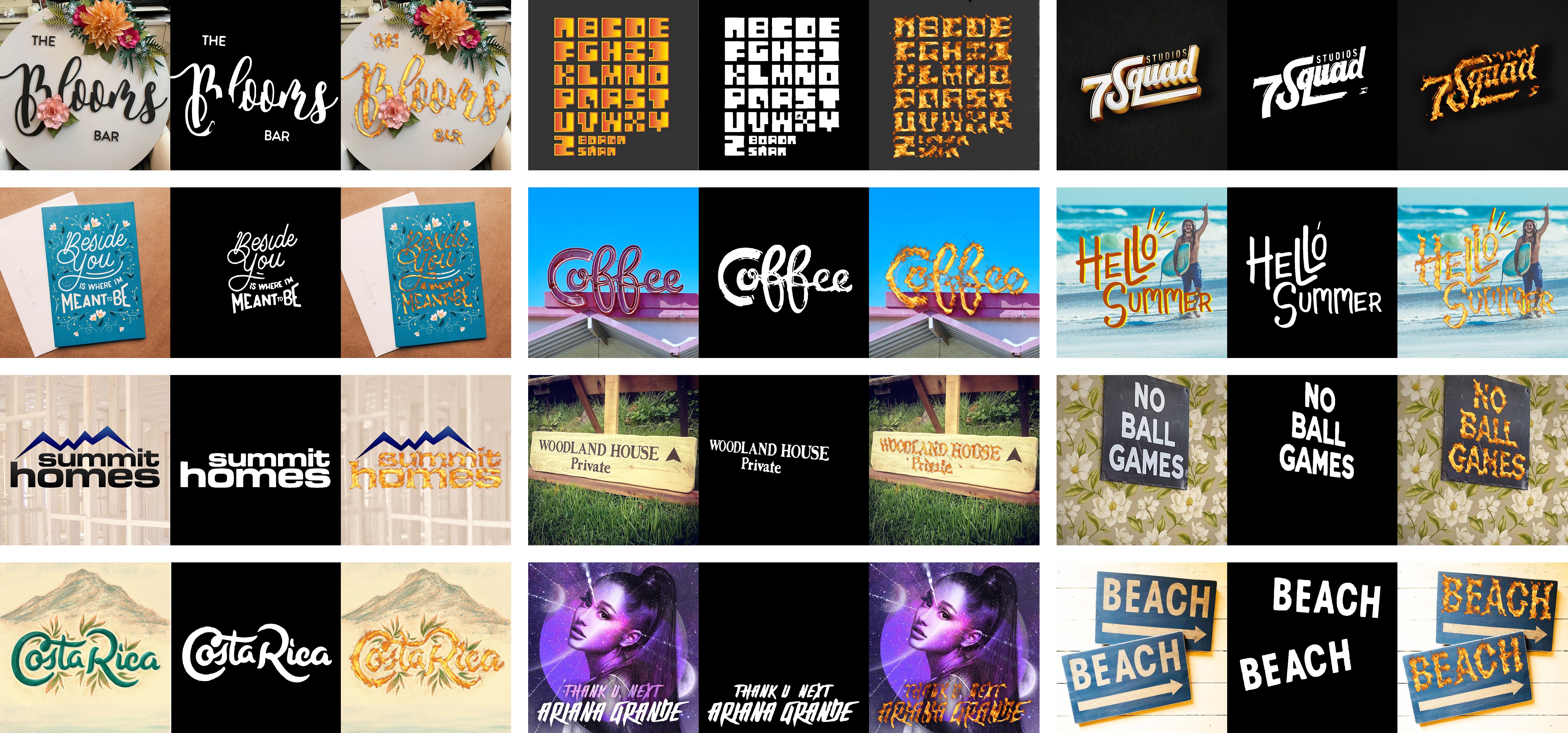}
    \vspace{0.3cm}
    \caption{Fire style}
    \label{fig:st_supp_fire}
    \vspace{0.5cm}
\end{subfigure}
\begin{subfigure}{.99\textwidth}
    \centering
    \includegraphics[width=.99\linewidth]{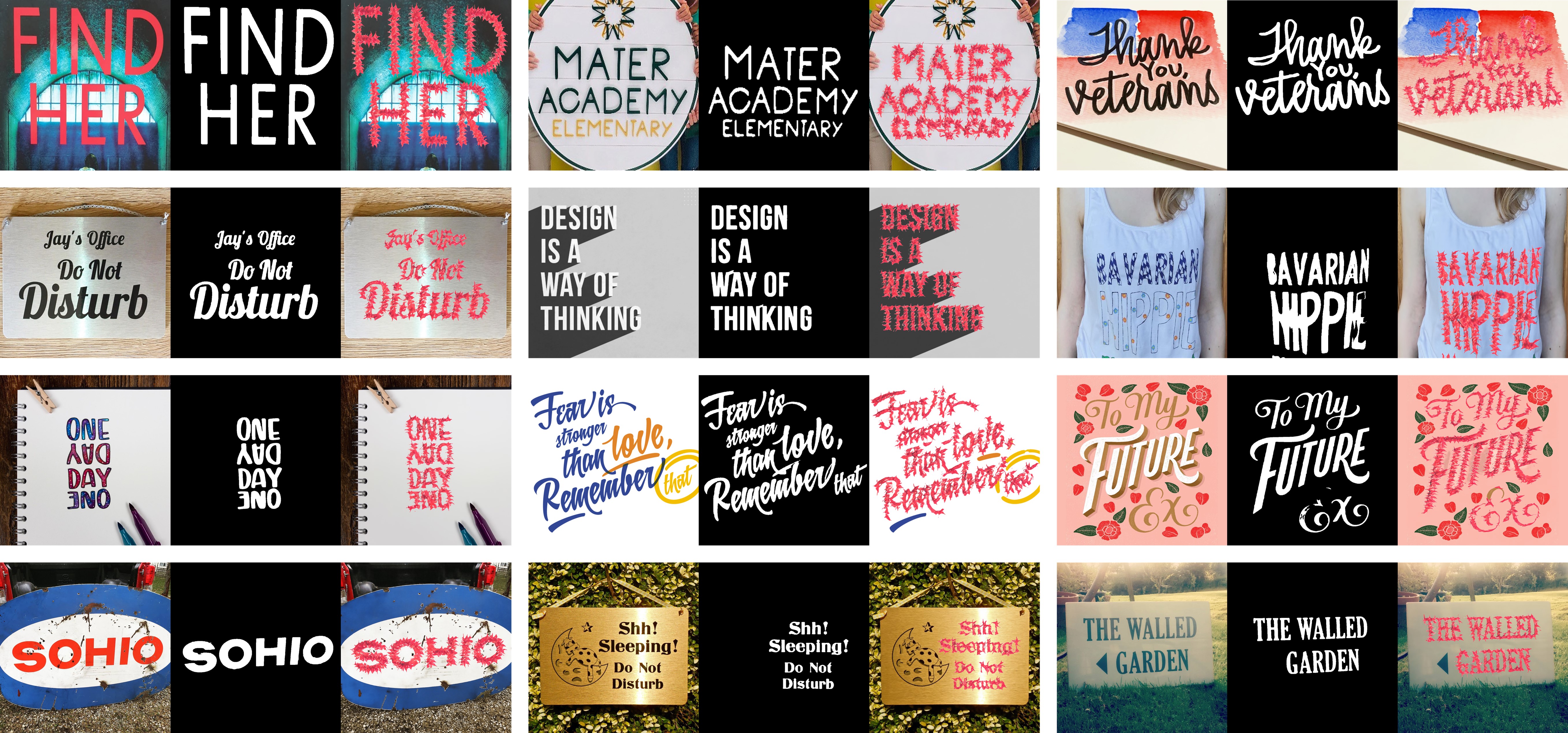}  
    \vspace{0.3cm}
    \caption{Maple style}
    \label{fig:st_supp_maple}
\end{subfigure}
\caption{Style transfer visualization using predicted text masks from our \modelname{} and text style transfer network Shape-Matching GAN~\cite{shapemgan}. For each sample, the original image, the predicted text mask, and the final result are shown from left to right.}
\label{fig:st_supp}
\end{figure*}

\begin{figure*}[t!]\ContinuedFloat
\centering
\begin{subfigure}{.99\textwidth}
    \centering
    \includegraphics[width=.99\textwidth]{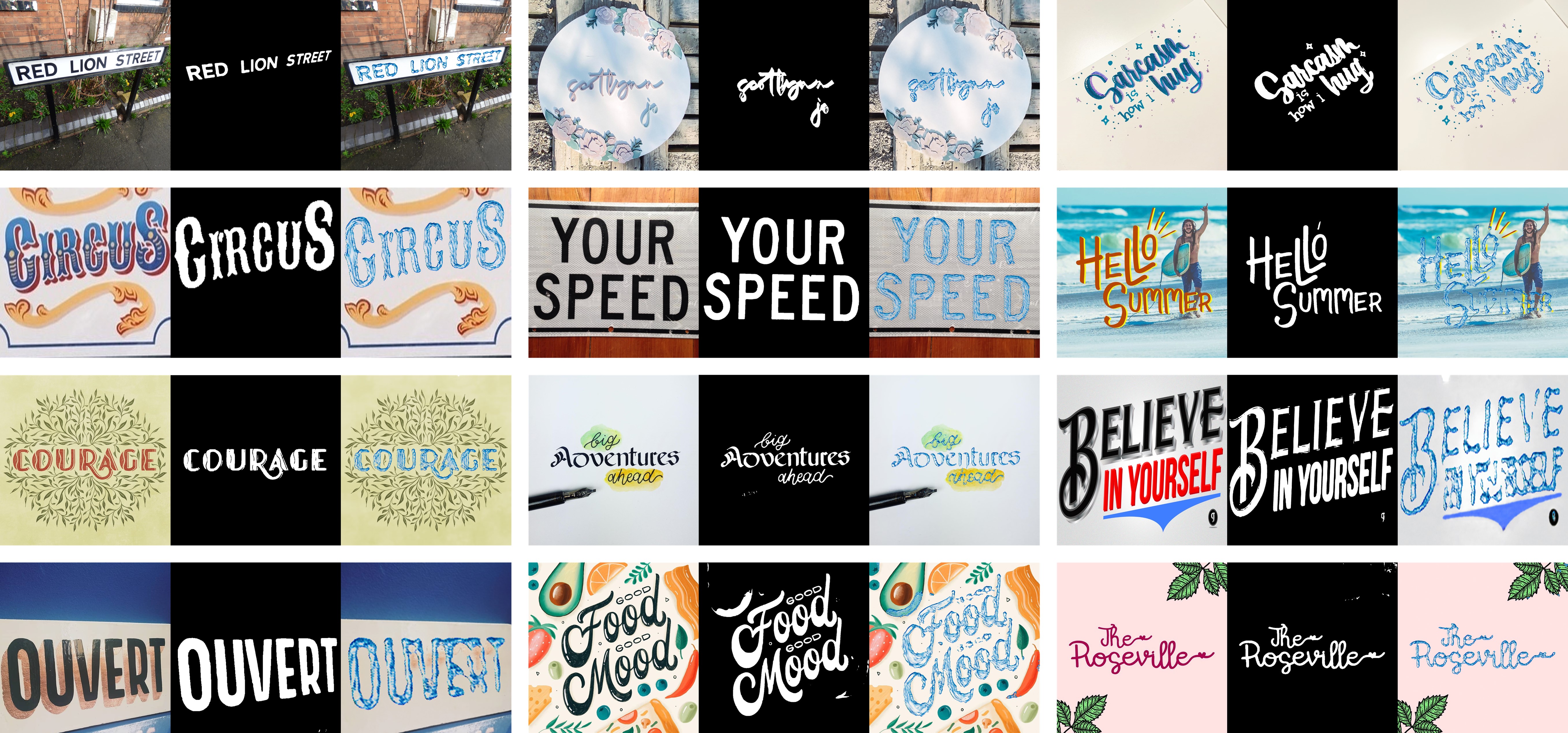}
    \vspace{0.3cm}
    \caption{Water style}
    \label{fig:st_supp_water}
\end{subfigure}
\vspace{0.3cm}
\caption{Style transfer visualization using predicted text masks from our \modelname{} and text style transfer network Shape-Matching GAN~\cite{shapemgan}. For each sample, the original image, the predicted text mask, and the final result are shown from left to right.}
\label{fig:st_supp}
\end{figure*}

\end{document}